\pdfoutput=1

\documentclass[11pt]{article}

\usepackage{EMNLP2023}

\usepackage{times}
\usepackage{latexsym}

\usepackage[T1]{fontenc}

\usepackage[utf8]{inputenc}

\usepackage{microtype}
\usepackage{enumerate}
\usepackage{enumitem}
\usepackage{multirow}
\usepackage{graphicx}
\usepackage{array}
\usepackage{booktabs}
\usepackage{inconsolata}
\usepackage{verbatim}
\usepackage{stmaryrd}
\usepackage{trimclip}
\makeatletter
\DeclareRobustCommand{\shortto}{%
  \mathrel{\mathpalette\short@to\relax}%
}
\newcommand{\short@to}[2]{%
  \mkern2mu
  \clipbox{{.5\width} 0 0 0}{$\m@th#1\vphantom{+}{\shortrightarrow}$}%
  }
\makeatother
\usepackage{amsmath,amsfonts,bm}

\newcommand{\Na}{({\em a})~}
\newcommand{\Nb}{({\em b})~}
\newcommand{\Nc}{({\em c})~}

\makeatletter   
\newcommand{\sveryshortarrow}[1][3pt]{\mathrel{%
    \vcenter{\hbox{\rule[-.5\fontdimen8\scriptfont3]
               {\scriptratio\dimexpr#1\relax}{\fontdimen8\scriptfont3}}}%
   \mkern-4mu\hbox{\let\f@size\sf@size\usefont{U}{lasy}{m}{n}\symbol{41}}}}
\makeatother

\def\eqref#1{equation~\ref{#1}}

\def\1{\bm{1}}

\def\m1{{\bm{1}}}

\DeclareMathAlphabet{\mathsfit}{\encodingdefault}{\sfdefault}{m}{sl}
\SetMathAlphabet{\mathsfit}{bold}{\encodingdefault}{\sfdefault}{bx}{n}

\def\gL{{\mathcal{L}}}

\def\gT{{\mathcal{T}}}

\def\sR{{\mathbb{R}}}

\def\sT{{\mathbb{T}}}

\usepackage[nameinlink]{cleveref}
\crefformat{section}{\S#2#1#3} 
\crefname{algorithm}{Alg.}{Algs.}
\crefformat{subsection}{\S#2#1#3}
\Crefname{equation}{Eq.}{Eqs.}
\Crefname{figure}{Fig.}{Figs.}
\newcommand{\Ni}{({\em i})~}
\newcommand{\Nii}{({\em ii})~}

\newcommand{\lsg}{\textsc{lsg}}
\newcommand{\dmea}{\textsc{dmea}}
\newcommand{\ie}{{\em i.e.,}}
\newcommand{\eg}{{\em e.g.,}}

\newcommand{\blue}[1]{\textcolor{blue}{#1}}

\usepackage{pifont}
\newcommand{\cmark}{\ding{51}}%
\newcommand{\xmark}{\ding{55}}%
\definecolor{applegreen}{rgb}{0.56, 0.8, 0.25}
\definecolor{mypurple}{rgb}{0.62,0.24,0.81}

\title{Lifelong Sequence Generation with 

Dynamic Module Expansion and Adaptation
}

\author{Chengwei Qin$^\clubsuit$, Chen Chen$^\clubsuit$, Shafiq Joty$^\clubsuit$$^\spadesuit$\\
$^\clubsuit$ Nanyang Technological University\\
$^\spadesuit$ Salesforce AI \\
\texttt{\{chengwei003@e.ntu, chen1436@e.ntu, srjoty@ntu\}.edu.sg}
}

\begin{document}
\maketitle
\begin{abstract}

Lifelong sequence generation (\lsg), a problem in continual learning, aims to continually train a model on a sequence of generation tasks to learn constantly emerging new generation patterns while avoiding the forgetting of previous knowledge. Existing \lsg\ methods mainly focus on maintaining old knowledge while paying little attention to knowledge transfer across tasks. In contrast, humans can better learn new tasks by leveraging previously acquired knowledge from similar tasks. Inspired by the learning paradigm of humans, we propose Dynamic Module Expansion and Adaptation (\dmea), {which enables the model to dynamically determine the  architecture for acquiring new knowledge based on task correlation} and select the most similar previous tasks to facilitate adaptation to new tasks. In addition, as the learning process can easily be biased towards the current task which might cause more severe forgetting of previously learned knowledge, we propose dynamic gradient scaling to balance the learning of the current task and replayed tasks. With extensive experiments, we demonstrate that \dmea\ can {consistently} outperform existing methods in different \lsg\ settings.
\end{abstract}

\section{Introduction} \label{sec:intro}

With the recent advancements in pre-trained language models (LMs), current sequence generation methods have achieved impressive performance on a variety of generation tasks \citep{radford2019language,raffel2020exploring,brown2020language,ouyang2022training,ding-etal-2023-gpt,zhao-etal-2023-verify,qin2023chatgpt}. Typically, these models are trained on a fixed corpus, assuming the underlying data distribution to be static  \citep{ham-etal-2020-end,el2021automatic}. However, real cognitive tasks are generally more complex involving changing contexts and dynamic environments. The ever-changing data distribution causes the models to face challenges in acquiring  new knowledge, while retaining the prior knowledge. Speaking about what is next for NLP, Kathleen McKeown in a recent  {interview} said: 
``Most models are static. But the world changes every minute, every second. Dealing with a dynamic world is a new area that's up and coming.''  \href{https://www.amazon.science/blog/acl-what-comes-next-for-natural-language-processing}{(Source)}

A potential solution is to formalize sequence generation as lifelong sequence generation or \lsg\ \citep{sun2020lamal}, where the model is expected to learn sequentially from a stream of generation tasks with potentially different data distributions. In such cases of distribution shift, the model might forget previously acquired knowledge upon learning new tasks, a phenomenon known as \emph{catastrophic forgetting} \citep{mccloskey1989catastrophic}. Previous \lsg\ methods \citep{mi-etal-2020-continual,sun2020lamal,madotto-etal-2021-continual} mainly explore  different ways to alleviate forgetting. Recently, \citet{zhang-etal-2022-continual} propose Adaptive Compositional Modules (ACM) which dynamically adds modules for new tasks depending on whether there are reusable previous modules, achieving SOTA performance on \lsg. 

Despite its effectiveness, ACM has several key limitations. First, {it mainly focuses on mitigating forgetting of previously acquired knowledge while paying little attention to transferring learned knowledge to new tasks which is as important for continual learning as preventing forgetting \citep{ke2020continual}}. In fact, a hallmark of human intelligence is that humans can better learn new tasks by leveraging previously acquired knowledge from similar tasks \citep{lake_ullman_tenenbaum_gershman_2017,qin-etal-2023-learning}. They can not only determine whether previously acquired skills are sufficient to solve a new task, but also exploit the most similar learned skills to facilitate the learning of the task; see \Cref{sec:human_learning} for an illustration. Second, ACM does not consider the correlation between learned tasks and the new task when adding modules, which might hinder finding the optimal architecture (case study in \Cref{sec:learned_model_architecture}). Finally, the learning process in ACM {can be biased towards the new task as the gradient norm of the new task on reused modules is typically much larger than that of replayed tasks, which may affect previously acquired knowledge; {see \Cref{sec:large_gradient} for an explanation}.}

Inspired by the learning paradigm of humans and to address the above limitations of ACM, in this work we propose Dynamic Module\footnote{Following \citet{zhang-etal-2022-continual}, we use an Adapter \citep{houlsby2019parameter} as the insertable module.} Expansion and Adaptation (\dmea). {We divide the learning process of a new task into three stages: expansion, selection and adaptation. In the expansion stage, \dmea\ determines whether to reuse modules of previous tasks or insert new modules for learning novel knowledge. Inspired by \citet{zhang-etal-2022-continual}, it utilizes differentiable architecture search \citep{liu2018darts} to enable the model to dynamically determine the architecture for solving the new task. {The learnable coefficients in architecture search are initialized based on the cosine similarity of word frequency distributions between learned tasks and the new task, aiming to discover the optimal model architecture. After searching, the module with the largest coefficient in every layer is chosen for the new task.} In the selection stage, \dmea\ selects the top-$K$ most similar previous tasks through input subspace \citep{lin2022trgp}. Finally, in the adaptation stage, it utilizes the selected similar tasks to facilitate adaptation to the new task.} The output of selected similar tasks is fused with that of the new task using learnable coefficients in every transformer layer to enable forward knowledge transfer. This is indeed an instance of mixture-of-experts  \citep{masoudnia2014mixture}. 

In addition, when the model learns a new task, \dmea\ also incorporates pseudo-sample replay \citep{sun2020lamal} to further mitigate catastrophic forgetting. To address the ``bias to the new task'' in the gradient update, we introduce dynamic gradient scaling to balance the learning of the new task and replayed tasks.   
To verify the effectiveness of \dmea, we conduct extensive experiments on various generation tasks in different \lsg\ settings. The empirical results show that \dmea\ can {consistently} outperform previous state-of-the-art baselines. 

In summary, our main contributions are:
\begin{itemize}[leftmargin=*,topsep=2pt,itemsep=2pt,parsep=0pt]
    \item {To the best of our knowledge, we are the first to explore solving \lsg\ from the perspective of human learning. We propose \dmea, a novel method based on dynamic module expansion and adaptation, to alleviate catastrophic forgetting and facilitate knowledge transfer in \lsg.}
    
    \item With extensive experiments and analysis, we demonstrate the effectiveness of our method compared to existing ones in different \lsg\ settings.
\end{itemize}

\section{Related Work}

\noindent\textbf{Lifelong Learning} (LL) aims to continually learn knowledge from a sequence of tasks with different distributions. The goal is twofold: alleviate \emph{catastrophic forgetting} \citep{mccloskey1989catastrophic} of learned tasks, and facilitate knowledge transfer \citep{lopez2017gradient} across tasks. 

Catastrophic forgetting typically means that the model forgets previously acquired knowledge after learning new tasks. Prior LL methods mainly focus on mitigating this problem and can be divided into three categories. First, \emph{regularization-based} methods constrain the update of parameters that are important to learned tasks to retain previous knowledge \citep{kirkpatrick2017overcoming,li2017learning,zenke2017continual,ritter2018online}. Second, \emph{architecture-based} methods dynamically adjust the model architecture to acquire new information while preventing the forgetting of previously learned tasks \citep{rusu2016progressive,chen2015net2net,fernando2017pathnet,madotto-etal-2021-continual,zhang-etal-2022-continual,qin2022lfpt}. Finally, \emph{memory-based} methods keep a number of key samples from previous tasks in memory to alleviate forgetting \citep{rebuffi2017icarl,shin2017continual,chaudhry2018efficient,qin-joty-2022-continual}. The memory data can be either real examples \citep{han-etal-2020-continual} or generated by language models \citep{sun2020lamal}. 

More recently, researchers have considered exploring knowledge transfer in LL, \ie\ learning on a task can benefit from learning on another task by transferring related knowledge. This includes CTR \citep{ke2021achieving} and CUBER \citep{lin2022beyond}. Despite their effectiveness, these methods mainly focus on classification tasks, while generation tasks typically have more complex label space. Note that this line of research is different from transfer learning \citep{ruder-etal-2019-transfer}, which mainly focuses on exploring better ways to reuse learned knowledge which is usually static, \eg\ a frozen language model. In contrast, the acquired knowledge is continually accumulated in lifelong learning.

\noindent\textbf{Lifelong Sequence Generation} (\lsg) enables the model to learn sequentially from a stream of generation tasks. \citet{sun2020lamal} propose LAMOL which formalizes different types of tasks as question answering and utilizes pseudo-sample replay to alleviate forgetting. \citet{chuang-etal-2020-lifelong} further improve LAMOL by knowledge distillation \citep{hinton2015distilling}. AdapterCL \citep{madotto-etal-2021-continual} inserts task-specific modules into every transformer layer to learn new tasks while keeping the pre-trained LM and previous modules frozen. On the basis of AdapterCL, \citet{zhang-etal-2022-continual} introduce ACM which dynamically adds modules for learning new tasks depending on whether there are reusable previously inserted modules. Though ACM can enable knowledge transfer to some extent via module sharing, there is no explicit mechanism to encourage knowledge transfer across tasks, {a common phenomenon of human learning.}

\noindent\textbf{Summary.} Existing work in \lsg\ mainly focuses on mitigating the catastrophic forgetting of previously learned knowledge while paying little attention to knowledge transfer across tasks. In contrast to these lines of work, we aim to explicitly encourage forward knowledge transfer in \lsg\ inspired by the way humans learn \citep{lake_ullman_tenenbaum_gershman_2017}.
\section{Problem Formulation}

\lsg\ involves learning from a stream of sequence generation tasks $\sT = (\gT^1, ...,\gT^n)$, where every task $\mathcal{T}^i$ has its own training set $D_{\text{train}}^i$, validation set $D_{\text{valid}}^i$, and test set $D_{\text{test}}^i$. Every dataset $D$ contains a set of examples $\{ (X_j,Y_j) \}_{j=1}^{|D|}$, where $X_j$ and $Y_j$ denote the input and output texts, respectively. At time step $k$, the model is trained on the training set $D_{\text{train}}^k$ of task $\mathcal{T}^k$ and has no access to real samples of previously learned tasks.

After the training on $D_{\text{train}}^k$, the model is expected to perform well on all the tasks learned so far, \ie\ $\gT^1,...,\gT^k$, and will be evaluated on the test set $D_{\text{test}}^i$ of each task $\mathcal{T}^i (1 \le i \le k) $ with corresponding evaluation metrics separately. Therefore, to achieve the goal of \lsg, the model is required to alleviate the forgetting of acquired knowledge and better learn new patterns through possible forward knowledge transfer.

\subsection{Data Format}  \label{sec:format}

Given an input-output text pair $(X, Y)$ for a task, the model learns to decode the output text $Y$ after reading the input $X$. Following \citet{zhang-etal-2022-continual}, a natural language question $Q$ describing the purpose of each task (task instruction) is inserted after the input to form a triple $(X, Q, Y)$; see \Cref{sec:ins_example} for an example. To learn a new task, the model is optimized to decode $Y$ given $X$ and $Q$. Denoting the concatenation of $X, Q$ and $Y$ as $A$, the autoregressive training objective is:
\begin{align}
\gL_{\text{task}} = - \sum_{j=m+1}^{n} \log p_{\theta} (A_j| A_{<j}) \label{eq:new_obj}
\end{align}
where $n$ is the total number of tokens in $A$ and $(A_1, ..., A_m)$ is the concatenation of $X$ and $Q$, and $\theta$ denotes the model parameters.

\section{Methodology}

Inspired by how humans learn a new task (\Cref{fig:human}), {\dmea\ divides the learning process into three stages. The \emph{expansion} stage (\Cref{sec:expansion}) first determines the model architecture dynamically. The \emph{selection} stage (\Cref{sec:selection}) then selects the top-K most similar previous tasks which are utilized in the final \emph{adaptation} stage (\Cref{sec:adaptation}) to facilitate adaptation to the new task.} We also employ pseudo-sample replay along with a dynamic gradient scaling method to balance the learning of the new and replayed tasks.

\begin{figure}[t]
    \centering
    \includegraphics[width=0.45\textwidth]{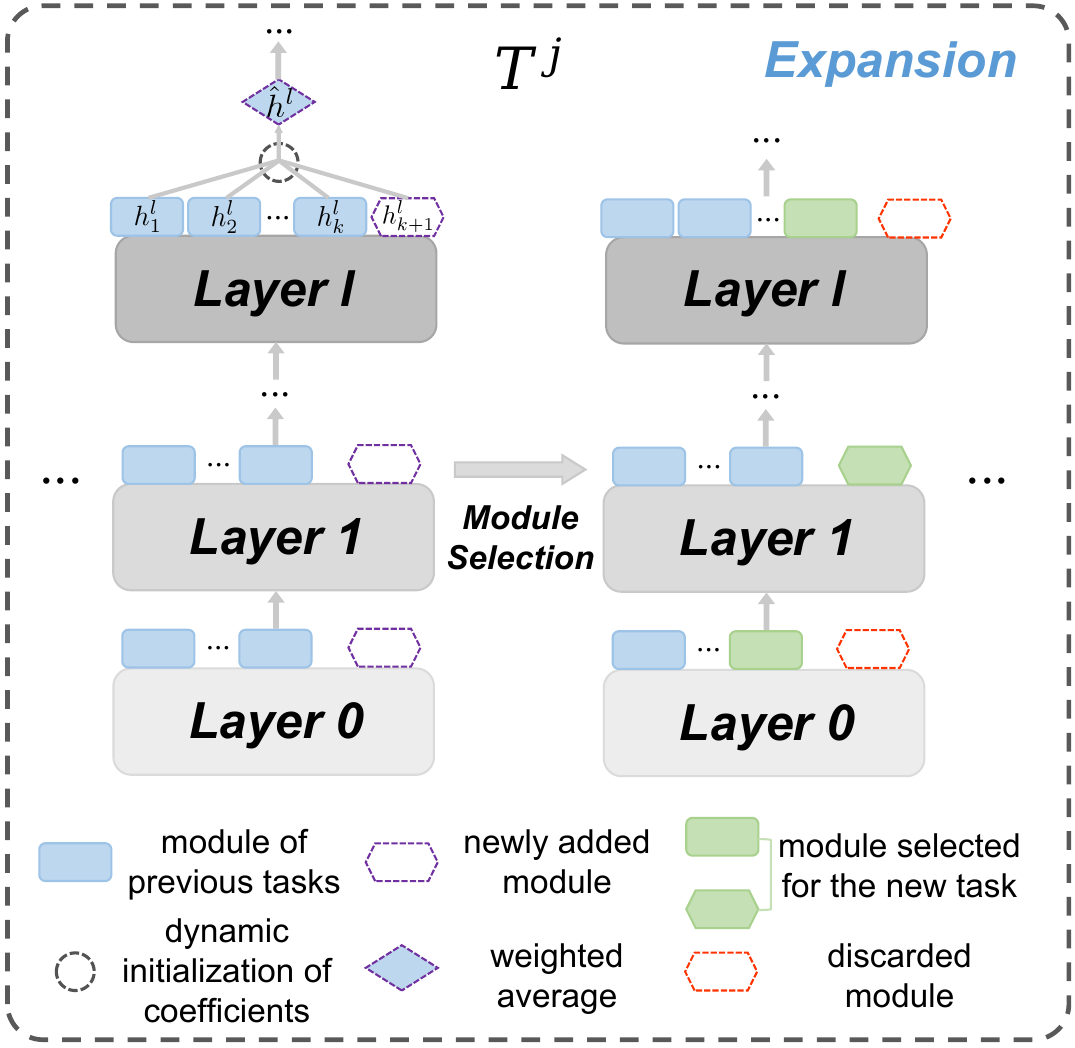}
    \caption{In the \textbf{expansion} stage, after inserting a new module (\textcolor{mypurple}{purple} dashed hexagon) into each layer, \dmea\ dynamically determines the architecture by differentiable architecture search. Specifically, the outputs of all modules in the same layer are fused through dynamically initialized learnable coefficients. The weighted average is then passed to the next layer of the model for learning. Note that only newly added modules (\textcolor{mypurple}{purple} dashed hexagons) are learnable modules in this stage. After several epochs of training, the module with the \emph{largest} coefficient in every layer (\textcolor{applegreen}{green} polygon) is selected for the new task. The selected module can be either a previous module (rectangle) or the newly added one (hexagon). Finally, newly added modules that are not selected (\textcolor{red}{red} dashed hexagons) will be discarded.}
    \vspace{-1em}
    \label{fig:method_expansion}
\end{figure}

\subsection{Expansion Stage}  \label{sec:expansion}
Humans are able to determine whether previously acquired skills are sufficient to solve a new task. Our method \dmea\ aims to mimic this learning process in the expansion stage. It can dynamically decide whether to reuse modules of previous tasks or insert a new module in every transformer layer to learn novel knowledge. Inspired by \citet{zhang-etal-2022-continual}, we utilize differentiable architecture search \citep{liu2018darts} to achieve this goal. 

Specifically, assuming that there are $k$ modules (\ie\ Adapter \citep{houlsby2019parameter}) $\{m_1^l,...,m_k^l\}$ in layer $l$ of the transformer model  before learning a new task $\mathcal{T}^j$, we temporarily insert a new module $m_{k+1}^l$ into this layer at the beginning of the expansion stage. For each forward pass, after calculating the output $h_t^l$ of every module $m_t^l$ in the layer separately, we fuse all outputs $\{h_1^l,...,h_{k+1}^l\}$ through learnable coefficients $\{\lambda_1^l,...,\lambda_{k+1}^l\}$ as follows.
\begin{align}
\hat h^l  = \sum_{t=1}^{k+1} \frac{e^{\lambda_t^l}}{\sum_{s=1}^{k+1} e^{\lambda_s^l}} h_t^l \label{eq:expansion}
\end{align}
The weighted average $\hat h^l$ is then passed to the next part of the model for learning. After training the model on $D_{\text{train}}^j$ for several epochs using $\gL_{\text{train}}$ (defined in \Cref{sec:adaptation}), we select the module with the \textbf{largest} coefficient in every layer for the new task $\mathcal{T}^j$. 

{Different from \citet{zhang-etal-2022-continual} which initialize $\{\lambda_1^l,...,\lambda_{k+1}^l\}$ with predefined hyperparameters, we propose to dynamically initialize learnable coefficients based on the correlation between the learned tasks $\mathcal{T}^1,...,\mathcal{T}^{j-1}$ and new task $\mathcal{T}^j$. Denoting the word frequency distribution of $\mathcal{T}^i$ as $f^i$ and all previous tasks sharing the module $m_t^l$ as $\mathcal{Z}_t^l$, the learnable coefficient $\lambda_t^l$ is initialized as:
\begin{equation}
\lambda_t^l=\left\{
\begin{aligned}
& \max_{\mathcal{T}^i \in \mathcal{Z}_t^l} \text{cos}(f^i, f^{k+1}), & 1 \le t \leq k \\
& \min_{1 \le i \leq k} \lambda_i^l, & t = k + 1
\end{aligned}
\right.
\end{equation}
where $\text{cos}$ is the cosine similarity function and $f^i$ is calculated based on the training set $D_{\text{train}}^i$. In this way, a previous module shared by tasks with higher word frequency distribution similarity to the new task has a larger initial coefficient, increasing the tendency to reuse it. In addition, the coefficient $\lambda_{k+1}^l$ of the newly added module $m_{k+1}^l$ is initialized to the minimum value of the initial coefficients $\{\lambda_1^l,...,\lambda_{k}^l\}$ of previously added modules $\{m_1^l,...,m_k^l\}$ to encourage module reuse.}

{The selected module in layer $l$ can be either from previous modules $\{m_1^l,...,m_k^l\}$  or the newly added one $m_{k+1}^l$ and will be tuned in the adaptation stage to accommodate new knowledge.} We then discard newly added modules that are not selected. Note that only newly added modules and coefficients are learnable in the expansion stage; the pre-trained LM and previous modules are kept frozen.

\subsection{Selection Stage}  \label{sec:selection}

\begin{figure}[t]
    \centering
    \includegraphics[width=0.45\textwidth]{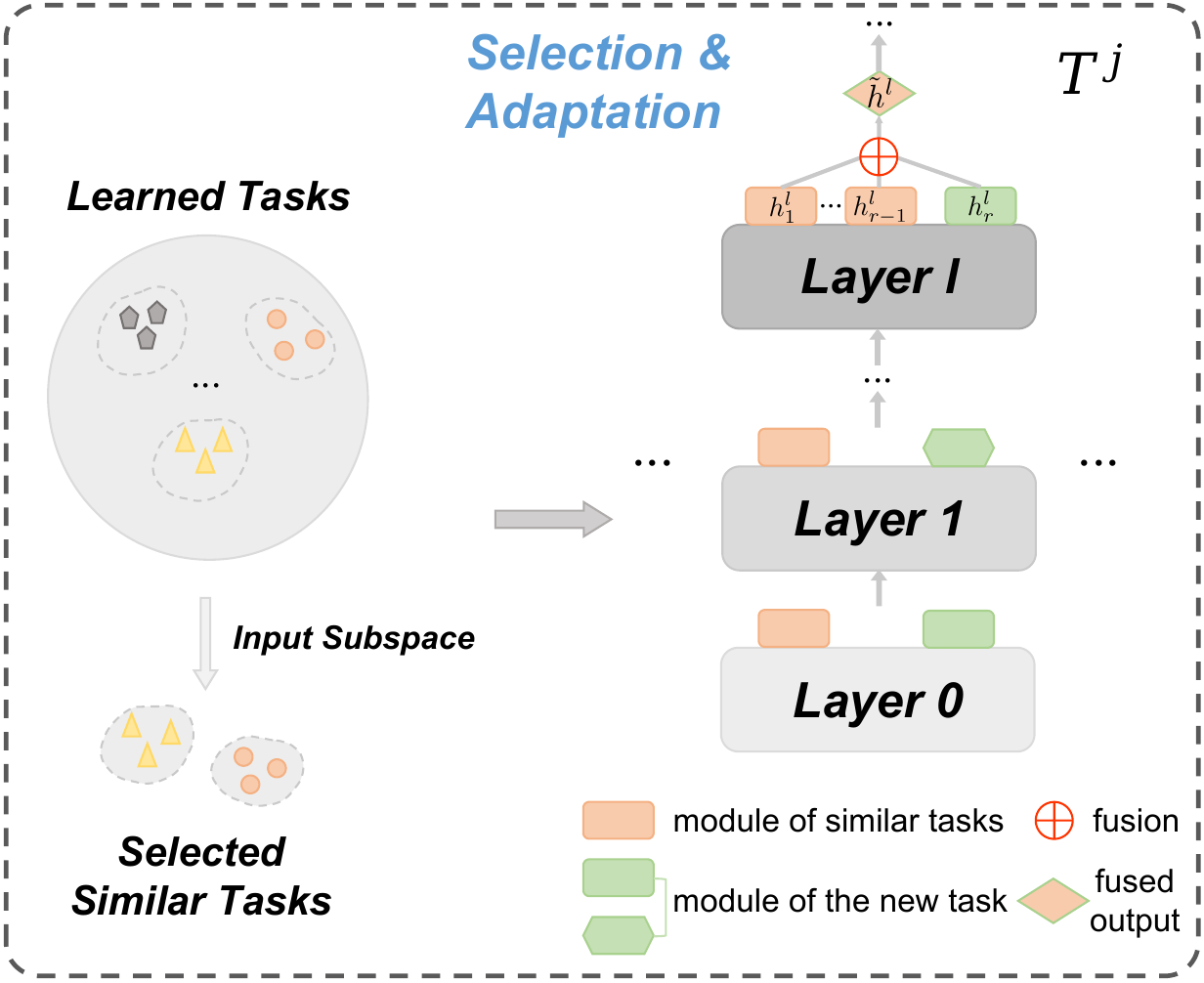}
    \caption{In the \textbf{selection} stage, \dmea\ selects the top-$K$ most similar previous tasks through input subspace to facilitate adaptation to a new task. During \textbf{adaptation}, the output of the selected similar tasks is fused with that of the new task in every layer to enable forward knowledge transfer. Note that only modules selected for the new task (\textcolor{applegreen}{green} polygons) are learnable modules in the adaptation stage. In addition, \dmea\ introduces dynamic gradient scaling to balance the learning of the new task and replayed tasks.}
    \vspace{-1em}
    \label{fig:method_selection_expansion}
\end{figure}

As humans, we can better acquire new knowledge by recognizing and utilizing knowledge from previously learned tasks that are  similar \citep{lake_ullman_tenenbaum_gershman_2017}. Based on the observation that the norm of one task's gradient projection onto the subspace of another task can characterize the correlation between them when the model architecture is static \citep{lin2022trgp}, we further extend it to dynamic modules. Specifically, we obtain the input subspace of each task using modules of it and select the top-$K$ most similar previous tasks by input subspace similarity to facilitate adaptation to the new task $\mathcal{T}^j$. {The model architecture induced from the expansion stage is used for selection and adaptation.}

Similar to \citet{lin2022trgp}, we adopt Singular Value Decomposition (SVD) to obtain the input subspace of each task. After training the model on $D_{\text{train}}^j$ for several epochs in the expansion stage, we randomly select $n$ samples $\{X_1,...,X_n\}$ from $D_{\text{train}}^j$ and obtain their representations $\{\bm{X}_1,...,\bm{X}_n\} \in \sR^m$ by forward-propagating them through the network. We use the final-layer representation of the last non-padding token in the input as the sample representation.

After obtaining the representation matrix $\bm{R}^j = [\bm{X}_1,...,\bm{X}_n] \in \sR^{m \times n}$ for task $\mathcal{T}^j$, we apply SVD to $\bm{R}^j$, \ie\ $\bm{R}^j = \bm{U}^j \bm{\Sigma}^j (\bm{V}^j)^{'}$, where $\bm{U}^j=[\bm{u}^j_1,...,\bm{u}^j_m] \in \sR^{m \times m}$ is composed of left-singular vectors $\bm{u}^j_i$, $\bm{\Sigma}^j  \in \sR^{m \times n}$ is a rectangular diagonal matrix with singular values on the diagonal, and $\bm{V}^j=[\bm{v}^j_1,...,\bm{v}^j_n] \in \sR^{n \times n}$ is composed of right-singular vectors $\bm{v}^j_i$. To obtain the input subspace $\bm{S}^j$ of $\mathcal{T}^j$, we select the first $k$ left-singular vectors in $\bm{U}^j$ to form the bases $\bm{B}^j = [\bm{u}^j_1,...,\bm{u}^j_k]$ for $\bm{S}^j$, where $k$ is determined by the requirement: $||\bm{R}^j_k||^2_F \ge \epsilon^j ||\bm{R}^j||^2_F$ with $\bm{R}^j_k$ being the $k$-rank approximation of $\bm{R}^j$, $F$ being the Frobenius norm, and $\epsilon^j$ being a predefined threshold.

For the new task $\mathcal{T}^j$, the norm of its subspace projection onto the subspace of a previously learned task  $\mathcal{T}^i$ could characterize the similarity $\bm{Q}_{j,i}$ between these two tasks. More formally, 
\begin{align}
\bm{Q}_{j,i} = \frac{||\text{Proj}_{\bm{S}^i}(\bm{S}^j)||_2}{||\bm{B}^j||_2}
\end{align}
where $\text{Proj}_{\bm{S}^i}(\bm{S}^j) = \bm{B}^j \bm{B}^i (\bm{B}^i)^{'}$ denotes the subspace projection. After getting the similarity scores $\bm{Q}_{j,i}, 1 \le i < j$ of all previous tasks, we pick $K$ tasks $\sT^{\text{sim}} = (\gT^{1},...,\gT^{K})$ with the top-$K$ highest scores to facilitate adaptation to the new task $\mathcal{T}^j$. 
 
\begin{table*}[t]
\centering
\small
\scalebox{0.83}{
\begin{tabular}{ll| p{1.3cm}<{\centering\arraybackslash} p{1.3cm}<{\centering\arraybackslash} p{1.3cm}<{\centering\arraybackslash} p{1.3cm}<{\centering\arraybackslash}| p{1.3cm}<{\centering\arraybackslash} p{1.3cm} <{\centering\arraybackslash} p{1.3cm} <{\centering\arraybackslash} p{1.3cm}<{\centering\arraybackslash} | p{1.3cm}<{\centering\arraybackslash}}
\toprule
\multicolumn{2}{l|}{\textbf{Methods}} &
  \textbf{Finetune} &
  \textbf{EWC} &
  \textbf{LAMOL} &
  \textbf{Metac} &
  \textbf{\begin{tabular}[c]{@{}l@{}}Adapter\\ +LAMOL\end{tabular}} &
  \textbf{\begin{tabular}[c]{@{}l@{}}AdapterCL\end{tabular}} &
  \textbf{\begin{tabular}[c]{@{}l@{}}ACM \end{tabular}} &
  \textbf{\dmea} &
  \textbf{\begin{tabular}[c]{@{}l@{}}MTL \end{tabular}} \\ \midrule
  \multicolumn{2}{l|}{Tune Whole Model?} & \cmark & \cmark & \cmark & \cmark & \xmark & \xmark & \xmark & \xmark & \cmark \\ \midrule
  \multicolumn{2}{l|}{Pseudo-sample Replay?} & \xmark & \xmark & \cmark & \cmark & \cmark & \xmark & \cmark & \cmark & ---  \\ \midrule
\multirow{4}{*}{Similar Tasks}                            & \# 1   &$42.9_{\pm 0.2}$ & $56.6_{\pm 0.1}$ & $65.5_{\pm 0.3}$ & $65.2_{\pm 0.4}$ & $64.5_{\pm 0.2}$ & $63.4_{\pm 0.3}$ & $64.8_{\pm 0.3}$ & $\textbf{65.8}_{\pm 0.2}$  & $67.1_{\pm 0.2}$ \\
                                                          & \# 2   &$51.8_{\pm 0.1}$ & $61.0_{\pm 0.2}$ & $65.2_{\pm 0.4}$ & $65.0_{\pm 0.2}$  & $64.6_{\pm 0.4}$ & $63.4_{\pm 0.3}$ & $64.9_{\pm 0.2}$ & $\textbf{65.6}_{\pm 0.2}$ & --- \\
                                                          & \# 3   &$45.2_{\pm 0.2}$ & $57.8_{\pm 0.1}$ & $\textbf{65.7}_{\pm 0.2}$ & $65.4_{\pm 0.4}$ & $64.2_{\pm 0.3}$ & $63.4_{\pm 0.3}$ & $64.5_{\pm 0.2}$ & $65.5_{\pm 0.1}$ & ---  \\
                                                          & \# 4   &$31.4_{\pm 0.4}$ & $46.6_{\pm 0.3}$ & $65.6_{\pm 0.3}$ & $65.5_{\pm 0.1}$ & $65.1_{\pm 0.1}$ & $63.4_{\pm 0.3}$ & $65.4_{\pm 0.3}$ & $\textbf{66.2}_{\pm 0.3}$ & --- \\ \hline
\multicolumn{2}{l|}{Average}                                       & $42.9_{\pm 8.5}$ & $55.5_{\pm 6.2}$ & $65.5_{\pm 0.2}$ & $65.3_{\pm 0.2}$ & $64.6_{\pm 0.4}$ & $63.4_{\pm 0.0}$ & $64.9_{\pm 0.4}$ & $\textbf{65.8}_{\pm 0.3}$ & $67.1_{\pm 0.0}$ \\
\midrule
\multirow{4}{*}{Random Tasks}                            & \# 1   &$33.8_{\pm 0.4}$ & $37.6_{\pm 0.5}$ & $55.7_{\pm 0.4}$ & $55.9_{\pm 0.3}$ & $53.2_{\pm 0.2}$ & $56.1_{\pm 0.5}$ & $56.7_{\pm 0.3}$ & $\textbf{57.5}_{\pm 0.3}$ & $59.7_{\pm 0.2}$ \\
                                                         & \# 2   & $33.1_{\pm 0.3}$ & $38.4_{\pm 0.4}$ & $62.6_{\pm 0.2}$ & $62.4_{\pm 0.4}$ & $61.8_{\pm 0.3}$ & $64.2_{\pm 0.2}$ & $64.9_{\pm 0.4}$ & $\textbf{65.6}_{\pm 0.3}$ & $67.5_{\pm 0.1}$  \\
                                                         & \# 3   &$25.7_{\pm 0.1}$ & $43.2_{\pm 0.3}$ & $54.8_{\pm 0.1}$ & $55.4_{\pm 0.2}$ & $53.6_{\pm 0.3}$ & $55.6_{\pm 0.4}$ & $56.3_{\pm 0.1}$ & $\textbf{57.3}_{\pm 0.2}$  & $60.4_{\pm 0.3}$\\
                                                         & \# 4   &$34.2_{\pm 0.3}$ & $48.9_{\pm 0.1}$ & $64.7_{\pm 0.4}$ & $65.3_{\pm 0.3}$  & $62.5_{\pm 0.1}$ & $65.4_{\pm 0.3}$ & $66.2_{\pm 0.2}$ & $\textbf{67.4}_{\pm 0.1}$ & $69.8_{\pm 0.1}$  \\ \hline
\multicolumn{2}{l|}{Average}                                      &$31.7_{\pm 4.0}$ & $42.0_{\pm 5.2}$ & $59.5_{\pm 4.9}$ & $59.8_{\pm 4.9}$ & $57.8_{\pm 5.1}$ & $60.3_{\pm 5.2}$ & $61.0_{\pm 5.3}$ & $\textbf{62.0}_{\pm 5.3}$ & $64.4_{\pm 5.1}$ \\ \bottomrule
\end{tabular}
}
\caption{\label{table-main-res}
The average performance score for each task sequence after learning all tasks. \textbf{Bold} indicates the best score. `MTL' stands for `multi-task learning', serving as the \emph{upper bound} for \lsg. {In each scenario, \dmea\ is significantly better than ACM with $p$-value $<0.05$ (paired t-test).} Note that while LAMOL and Metac are not directly comparable to other adapter-based methods as their learnable parameters are orders of magnitude larger, \dmea\ still outperforms them in most cases. The comparison of learnable parameters and computational resources between ACM and \dmea\ is reported in \Cref{sec:comparison_acm_dmea}.
}
\vspace{-1em}
\end{table*}
\subsection{Adaptation Stage}  \label{sec:adaptation}

For adaptation to $\mathcal{T}^j$, assume that $\sT^{\text{all}} = (\gT^{1},...,\gT^{K},\mathcal{T}^j)$ contains a total of $r$ modules $\{m_1^l,...,m_r^l\}$ in layer $l$. During the training on $D_{\text{train}}^j$ using $\gL_{\text{train}}$ (see \Cref{eq:whole_loss}), for each sample in $D_{\text{train}}^j$, we fuse the output $h_s^l$ of each module $m_s^l \in \{m_1^l,...,m_r^l\}$  by learnable coefficients $\{\alpha_1^l,...,\alpha_{r}^l\}$ to enable \emph{forward knowledge transfer}:
\begin{align}
\tilde h^l  = \sum_{s=1}^{r} \frac{e^{\alpha_s^l}}{\sum_{u=1}^{r} e^{\alpha_u^l}} h_s^l \label{eq:adaptation}
\end{align}
{The learnable coefficients $\{\alpha_1^l,...,\alpha_{r}^l\}$ are equally initialized to $1.0$.} Similar to the expansion stage, the fused output $\tilde h^l$ is passed to the next part of the model for learning. After training, the learnable coefficients will be saved for inference. Note that we only tune modules selected in the expansion stage (can be modules of previous tasks or newly added modules)
and learnable coefficients while keeping the pre-trained language model and other modules frozen.

As there is no saved real sample of previously learned tasks when the model adapts to a new task, we also incorporate pseudo-sample replay \citep{sun2020lamal} to alleviate the forgetting of acquired knowledge.  
We achieve this by simultaneously training the model as a task solver ($\gL_{\text{task}}$ in \Cref{sec:format}) and as a data generator. When training as a data generator, the model learns to generate the triple $(X, Q, Y)$ given a task-specific generation token $G$ as input. Then before learning a new task, the model can generate pseudo samples of previous tasks, which are combined with new data for training to mitigate forgetting. Denoting the concatenation of $G, X, Q$ and $Y$ as $A^{'}$, the data generation loss is expressed as:
\begin{align}
\gL_{\text{data}} = - \sum_{i=2}^{m} \log p_{\theta} (A^{'}_i| A^{'}_{<i}) \label{eq:data_loss}
\end{align}
where $m$ is the total number of tokens in $A^{'}$. The overall loss that \dmea\ optimizes for adapting to a new task is:
\begin{align}
\gL_{\text{train}} = \gL_{\text{task}} + \mu \gL_{\text{data}} \label{eq:whole_loss}
\end{align}
where $\mu$ is the weight of data generation loss.

After the expansion stage, if the new task reuses some modules of previously learned tasks, the model will generate some pseudo samples of these tasks and train the model using $\gL_{\text{train}}$ on the combination of new data and pseudo data. As the model has not seen new data before, the gradient norm of the new task on reused modules is much larger than that of replayed tasks. The learning process can easily be biased towards the new task which may affect previously acquired knowledge. 

Therefore, to balance the learning of the new task and replayed tasks, we introduce \emph{dynamic gradient scaling}. Specifically, assuming that the new task $\mathcal{T}^j$ reuses $s$ modules $\{m_1,...,m_s\}$ of a previous task $\mathcal{T}^i$ in all layers, we randomly select $q$ examples from $D_{\text{train}}^j$ and pseudo samples of $\mathcal{T}^i$ separately and forwards them through the model to obtain the gradient of $\mathcal{T}^j$ and $\mathcal{T}^i$ using $\gL_{\text{train}}$ with regard to reused modules $\{m_1,...,m_s\}$, denoted as $g^j$ and $g^i$, respectively. The dynamic scale factor $\eta_t^i$ is then calculated as:
\begin{align}
\eta_t^i = (\frac{||g^j||_2}{||g^i||_2} - 1) e^{-t} + 1 \label{eq:scale}
\end{align}
where $t$ is the number of completed training epochs. After dynamic gradient scaling, the total loss for jointly learning $\mathcal{T}^j$ and $\mathcal{T}^i$ is:
\begin{align}
\gL_{\text{total}} = \gL_{\text{train}}^j + \eta_t^i \gL_{\text{train}}^i \label{eq:total_dynamic}
\end{align}
Note that in the early stage of training, the value of $t$ is small. $\eta_t$ is greater than $1$ to balance the gradient of the new task $\mathcal{T}^j$ and the replayed task $\mathcal{T}^i$. When the model has seen enough new data in the late stage of training (no need to balance), $\eta_t$ is approximately equal to $1$ as the value of $t$ is large. 
\section{Experimental Setup} \label{sec:setup}

In this section, we first describe investigated tasks and then introduce methods compared in our work.

\subsection{Tasks}  \label{sec:tasks}

Four representative sequence generation tasks are investigated in our work: natural language generation, summarization, task-oriented dialogue and SQL query generation. Following \citet{zhang-etal-2022-continual}, we consider two different scenarios: \Ni \lsg\ on \emph{similar} tasks where the model learns a sequence of tasks of the same type but different domains, and \Nii \lsg\ on \emph{random} tasks where the model learns knowledge from different types of tasks. For \lsg\ on similar tasks, we use five different domains from two natural language generation datasets (RNNLG \citep{wen-etal-2015-semantically} and E2ENLG \citep{novikova-etal-2017-e2e}) to form the task sequences. We further incorporate summarization (CNNDM \citep{see2017get}), task-oriented dialogue (MultiWOZ \citep{budzianowski-etal-2018-multiwoz}) and SQL query generation (WikiSQL \citep{zhong2017seq2sql}) to form the task sequences for \lsg\ on random tasks. For each scenario, we randomly select four different orders\footnote{\citet{zhang-etal-2022-continual} sample data from the original set for data balance. To ensure a fair comparison among all methods, we resample new data for experiments.} (\Cref{sec:diff_task_orders}) and run experiments for every order five times with different random seeds (20 runs per scenario). For each order, we report the average of all learned tasks' performance scores following \citet{zhang-etal-2022-continual}; see \Cref{sec:task_metrics} for details of task-specific evaluation metrics.

\subsection{Methods Compared} \label{sec:methodscom}

Following \citet{zhang-etal-2022-continual}, we use GPT-2 \citep{radford2019language} as the backbone model and Adapter \citep{houlsby2019parameter} as the insertable module, and compare with the following methods:

\begin{itemize}[leftmargin=*,topsep=4pt,itemsep=4pt,parsep=0pt]

\item \textbf{Finetune} tunes the whole GPT-2 model only on the training data of the new task during the \lsg\ process.

\item \textbf{EWC} \citep{kirkpatrick2017overcoming} constrains the update of parameters that are important to previously learned tasks to alleviate forgetting.

\item \textbf{LAMOL} \citep{sun2020lamal} tunes the whole GPT-2 model with pseudo-sample replay.

\item \textbf{Metac-Adapt (Metac)} \citep{wang2023metacognitive} adapts LAMOL towards better semantic space for generating pseudo samples.

\item \textbf{Adapter+LAMOL} only inserts adapter modules for the first task and tunes these modules with pseudo-sample replay while keeping the backbone model frozen.

\item \textbf{AdapterCL} \citep{madotto-etal-2021-continual} inserts task-specific adapter modules for every new task while keeping the backbone model and previous modules frozen.

\item \textbf{ACM} \citep{zhang-etal-2022-continual}  dynamically adds adapter modules for new tasks depending on whether there are reusable previous modules to improve the performance and parameter efficiency of AdapterCL. It is the state-of-the-art on \lsg.

\end{itemize}

\section{Results and Analysis} \label{sec:exp_res}

\begin{table}[t]
\centering
\small
\setlength\tabcolsep{3pt}
\scalebox{0.90}{
\begin{tabular}{l|c|c}
\toprule
\multirow{1}{*}{\textbf{Method}} & \multicolumn{1}{c}{Similar Tasks}                    & \multicolumn{1}{|c}{Random Tasks}                    \\
\midrule

\dmea & \textbf{65.8} & \textbf{57.3} \\
\emph{w.o.} transfer & 64.9  & 56.5 \\
\emph{w.o.} scaling & 65.5  & 56.8  \\
\emph{w.o.} initialization & 65.4  & 57.0  \\
\bottomrule
\end{tabular}
}
\caption{ The average performance score for different ablations: (i) without forward knowledge transfer, (ii) without dynamic gradient scaling, and (iii) without dynamic initialization. All components improve the performance of our method.} 
\label{table:ablation-brief}
\end{table}

\subsection{Main Results} \label{sec:main_res}
\Cref{table-main-res} shows the average performance score for each task sequence after learning all tasks (see \Cref{sec:detail_performance} for the performance of each task). From the results, we can see that \dmea\ outperforms previous baselines in all \lsg\ settings, which demonstrates the superiority of our method. Note that while the learnable parameters of LAMOL are orders of magnitude larger, \dmea\ still achieves better performance than LAMOL in 7 out of 8 runs, showing its effectiveness in \lsg. 

Simply fine-tuning the model with new samples leads to poor performance due to catastrophic forgetting. Although EWC adopts Fisher information matrix to alleviate forgetting, its performance is still much worse than other memory-based baselines, indicating the importance of pseudo-sample replay. {When learning from a sequence of similar tasks, Adapter+LAMOL performs better than AdapterCL as AdapterCL applies parameter isolation to different tasks which might prevent positive knowledge transfer across tasks. However, this is not the case when learning from random tasks: AdapterCL achieves much better results than Adapter+LAMOL as AdapterCL can avoid catastrophic forgetting by assigning different learnable parameters to each task. The performance of ACM is superior to Adapter+LAMOL and AdapterCL in both scenarios, showing the effectiveness of its adaptive compositional architecture. However, ACM has no explicit mechanism to encourage forward knowledge transfer in \lsg, which is actually the human learning paradigm. Our proposed \dmea\ consistently outperforms ACM by dynamically leveraging previously acquired knowledge to facilitate adaptation to new tasks.}

\subsection{Ablation Study} \label{subsec:abl}
{We conduct several ablations to analyze the contribution of different components of \dmea. In particular, we investigate three variants of \dmea\: \Na without selecting similar previous tasks for forward knowledge transfer (\emph{w.o.} transfer), \Nb removing dynamic gradient scaling (\emph{w.o.} scaling), and \Nc without dynamically initializing learnable coefficients (\emph{w.o.} initialization).} For each scenario, \ie\ similar tasks or random tasks, we randomly pick one sequence for experiments. \Cref{table:ablation-brief} reports the average performance score after learning all tasks for different ablations. 

From the results, we can observe that all components contribute to the average performance. {Removing forward knowledge transfer leads to a significant performance drop in both scenarios, indicating that selecting top-$K$ most similar previous tasks can indeed discover and transfer useful learned knowledge to facilitate adaptation to the new task. The adoption of dynamic gradient scaling yields a moderate performance boost as it can balance the learning of the new task and replayed tasks to mitigate catastrophic forgetting. Dynamic initialization of learnable coefficients also facilitates performance improvement, demonstrating the effectiveness of leveraging the similarity of word frequency distributions between tasks.}

\begin{table}[t]
\centering
\small
\scalebox{0.90}{
\begin{tabular}{llccc}
\toprule
\multicolumn{2}{l}{\textbf{Time Step}} & \textbf{AdapterCL} & \textbf{ACM} & \textbf{\dmea}\\
\midrule
\multirow{4}{*}{Similar}                            & 2 & 55.8(+0.0)  &  56.0(+0.1)  & \textbf{56.3}(+\textbf{0.3})   \\  
                                                    & 3 & 58.6(+0.0)  &  59.1(+0.4)  & \textbf{59.5}(+\textbf{0.6})\\ 
                                                    & 4 & 61.2(+0.0)  &  62.5(+0.6)  & \textbf{63.2}(+\textbf{0.9})\\ 
                                                    & 5 & 63.4(+0.0)  &  64.8(+0.3)  & \textbf{65.8}(+\textbf{1.0})\\ 
\midrule
\multirow{4}{*}{Random}                            & 2  & 55.4(+0.0)  &  56.3(+0.9)  & \textbf{57.1}(+\textbf{2.1})\\ 
                                                    & 3 & 58.4(+0.0)  &  58.9(+0.3)  & \textbf{59.7}(+\textbf{1.3})\\
                                                    & 4 & 64.0(+0.0)  &  64.3(+0.7)  & \textbf{65.4}(+\textbf{1.4})\\ 
                                                    & 5 & 64.2(+0.0)  &  64.9(+0.6)  & \textbf{65.6}(+\textbf{1.1})\\ 
                                                    \bottomrule
\end{tabular}
}
\caption{\label{table-fkt}
The average performance score and forward knowledge transfer (FKT) of different methods at every time step. FKT is reported in parentheses.
}
\end{table}

\subsection{Further Analysis} \label{subsec:analysis}

\begin{table}[t]
\centering
\small
\setlength\tabcolsep{3pt}
\scalebox{0.90}{
\begin{tabular}{l|c|c}
\toprule
\multirow{1}{*}{\textbf{Metrics}} & \multicolumn{1}{c}{Similar Tasks}                    & \multicolumn{1}{|c}{Random Tasks}                    \\
\midrule

Input Subspace & \textbf{65.8} & \textbf{57.3} \\
Frequency & 65.3  & 56.9  \\
Representation & 65.2  & 56.9  \\
\emph{w.o.} transfer &  64.9 & 56.5 \\

\bottomrule
\end{tabular}
}
\caption{ The average performance score using different similarity metrics.} 
\label{table:similarity_metrics}
\end{table}

\paragraph{Quantify Forward Knowledge Transfer.} Following \citet{ke2020continual}, we define metrics quantifying forward knowledge transfer (FKT) at every time step $t$ as:
\begin{align}
 \text{FWT}  = \frac{1}{t-1} \sum_{i=2}^{t} R_{i,i} - \bar{d}_i. \label{eq:fwt} 
\end{align}
where $R_{i,j}$ is the performance score on $\mathcal{T}^j$ after learning $\mathcal{T}^i$ and {$\bar{d}_i$ refers to the performance of training $\mathcal{T}^i$ individually, which is actually the result of AdapterCL.} For each scenario, we randomly select one sequence for analysis and report the average performance score along with FKT at each step in \Cref{table-fkt}. From the results, we can see that \dmea\ consistently outperforms ACM in terms of the average performance score and FKT at all steps, demonstrating that \dmea\ can better facilitate positive knowledge transfer.

{\paragraph{Input Subspace vs. Other Similarity Metrics.} The ablation (\emph{w.o.} transfer) in \Cref{subsec:abl} demonstrates the importance of selecting similar learned tasks. To further investigate whether different similarity metrics influence the performance of \dmea, we conduct controlled experiments with two new metrics: \Na cosine similarity of word frequency distributions between different tasks (\emph{frequency}), and \Nb cosine similarity of the representations of selected samples from different tasks\footnote{{For a pair of tasks, we compute the cosine similarity for every representation pair and use the average as the similarity.}} (\emph{representation}). For each scenario, we use the same sequence as \Cref{subsec:abl}. From the results in \Cref{table:similarity_metrics}, we can observe that selecting similar previous tasks by input subspace consistently outperforms using other similarity metrics, demonstrating its superiority.}

\paragraph{Robustness to Module Type} To verify whether the performance gain of \dmea\ is consistent across different types of modules, we extend the experiments to prefix-tuning \citep{li-liang-2021-prefix} and LoRA \citep{hu2022lora}. We randomly pick four sequences for experiments and report the average result in \Cref{tab:diff_module_type}. we can see that \dmea\ still outperforms ACM when using other architecture as the insertable module, showing its robustness to module type.

\begin{table}[t]
\centering
\small
\setlength\tabcolsep{3pt}
\scalebox{0.95}{
\begin{tabular}{l|c|c}
\toprule
\multirow{1}{*}{\textbf{Module Type}} & \multicolumn{1}{c}{ACM}                    & \multicolumn{1}{|c}{\dmea}                    \\
\midrule
Prefix-tuning & 62.6  & \textbf{63.4}  \\
LoRA & 63.1 & \textbf{64.2} \\
\bottomrule
\end{tabular}
}
\caption{The average performance score of ACM and \dmea\ with different module types.} 
\label{tab:diff_module_type}
\end{table}

\begin{figure}[t]
    \centering
    \includegraphics[width=0.34\textwidth]{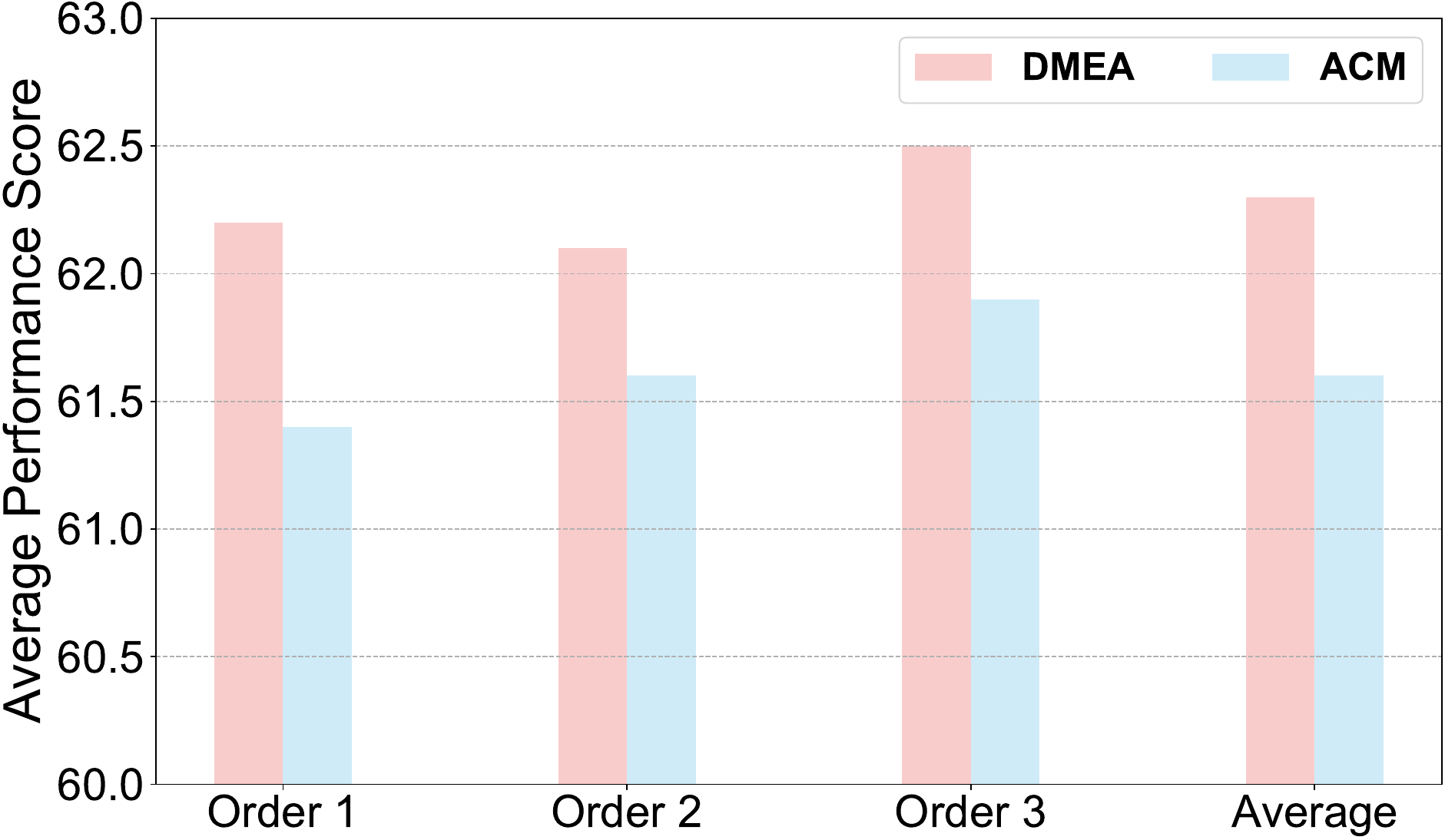}
    \caption{The average performance score for every order after learning all 8 tasks of the longer sequence.}
    \label{fig:longer_sequence}
\end{figure}

\paragraph{Longer Sequence.} As mentioned in \Cref{sec:tasks}, we mainly conduct experiments on sequences consisting of 5 tasks following \citet{zhang-etal-2022-continual}. To verify whether \dmea\ can still outperform the baselines when learning from a larger number of tasks, we further combine all tasks investigated in this work to form a longer sequence of 8 tasks. We evaluate ACM and \dmea\ on this longer sequence with 3 different orders and report the average performance score for each order after learning all tasks {in \Cref{fig:longer_sequence}. We can observe that \dmea\ is still superior to ACM when learning from longer sequences.}

\begin{figure}[t]
    \centering
    \includegraphics[width=0.45\textwidth]{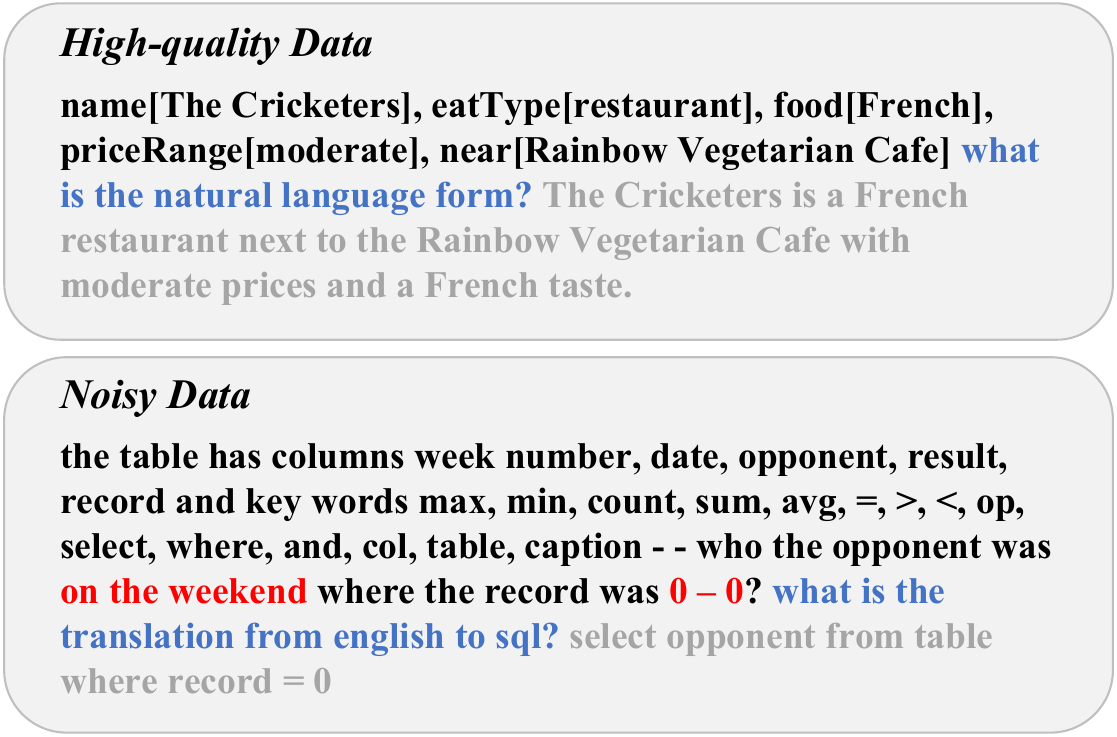}
    \caption{Some examples of generated pseudo data. We color the task instruction in \textcolor{blue}{blue} and output text in \textcolor{gray}{gray}. Missing/wrong information is colored in \textcolor{red}{red}.}
    \label{fig:pseudo}
\end{figure}

\paragraph{Quality of Pseudo Data} \Cref{fig:pseudo} shows several pseudo samples generated by \dmea. We can see that \dmea\ can indeed generate high-quality pseudo samples to mitigate the forgetting of previously learned knowledge. However, the generated pseudo data could also be noisy as shown at the bottom of the figure, which might hinder further performance improvement. 

\paragraph{Other Types of Tasks} To explore whether the performance gain of \dmea\ is consistent on other types of tasks, we further include three new tasks: sentiment analysis (SST \citep{socher-etal-2013-recursive}), semantic role labeling (SRL \citep{he-etal-2015-question}) and question answering (SQuAD \citep{rajpurkar-etal-2016-squad}). We randomly select two tasks from the original task set three times and combine them with new tasks to form three task sequences. From the results shown in \Cref{new_types}, we can observe that \dmea\ performs better than ACM on all sequences, showing its robustness to task types.

\paragraph{Different Pseudo-data Sampling Ratios} Following \citet{zhang-etal-2022-continual}, we set the pseudo-data sampling ratio to $0.2$. To validate whether different pseudo-data sampling rates influence the performance gain of \dmea, we conduct controlled experiments with sampling rates $\{0.05,0.1,0.4\}$. We randomly pick three sequences for experiments and report the performance comparison between ACM and \dmea\ in \Cref{tab:diff_pseudo_data_sampling_rate}. We can see that \dmea\ consistently outperforms ACM in all cases, demonstrating its effectiveness.

In addition, we show case studies of learned model architecture, model output, dynamic gradient scaling and task selection, generalization of dynamic initialization, and potential real-world applications in Appendix \ref{sec:learned_model_architecture} $\sim$ \ref{sec:real_world_application}, respectively.

\begin{table}[t]
    \centering
    \begin{tabular}{lcccc}
    \toprule
    \multicolumn{1}{l}{\multirow{2}*{\textbf{Method}}}& \multicolumn{3}{c}{\textbf{Sequence}} &
    \multicolumn{1}{c}{\multirow{2}*{\textbf{Average}}}\\
    \cmidrule{2-4}
    {} & (i) & (ii) & (iii) \\
    \midrule
    \multicolumn{1}{l}{ACM} {} & 68.4 & 63.6 & 71.8 & 67.9 \\
    \multicolumn{1}{l}{\dmea} {} & \textbf{69.5} & \textbf{64.4}  & \textbf{73.0} & \textbf{69.0} \\
    \bottomrule
    \end{tabular}
    \caption{The average performance score for every sequence after learning all new types of tasks. 
    }
    \label{new_types}
\end{table}

\begin{table}[t]
\centering
\small
\setlength\tabcolsep{3pt}

\begin{tabular}{l|c|c|c}
\toprule
\multirow{1}{*}{\textbf{Sampling Ratio}} & \multicolumn{1}{c}{0.05}                    & \multicolumn{1}{|c}{0.1} & \multicolumn{1}{|c}{0.4}                     \\
\midrule
ACM & 61.7  & 62.0 & 62.1 \\
\dmea & \textbf{62.5} & \textbf{63.1}  & \textbf{62.8} \\
\bottomrule
\end{tabular}
\caption{The average performance score of ACM and \dmea\ with different pseudo-data sampling ratios.} 
\label{tab:diff_pseudo_data_sampling_rate}
\end{table}
\section{Conclusion} \label{sec:conclusion}
In this work, we have introduced \dmea\ for lifelong sequence generation (\lsg). \dmea\ leverages task correlations to dynamically determine the suitable architecture required to acquire novel knowledge of a new task and selects the most similar previous tasks through input subspace to facilitate knowledge transfer. It uses pseudo-sample replay along with dynamic gradient scaling to balance the learning of the new task and replayed tasks to further alleviate forgetting. With extensive experiments and analysis we have shown that \dmea\ {consistently} outperforms previous methods in different \lsg\ settings. In the future, we would like to investigate ways to improve the quality of pseudo data and explore more metrics for task similarity.

\section*{Limitations}

Although effective, \dmea\ has couple of limitations: 

\begin{itemize}[leftmargin=*,topsep=2pt,itemsep=2pt,parsep=0pt]
    \item \dmea\ mainly focuses on the setting where every task has plenty of training samples. In contrast, humans can easily learn to perform new tasks with only few data, which is a hallmark of human intelligence. We leave how to explore lifelong sequence generation in few-shot settings as future work.
    \item \dmea\ does not consider machine translation, a sequence generation task that might involve vocabulary changes. One potential solution is to use multilingual pre-trained language models.
\end{itemize}

\bibliography{anthology,custom}

\begin{thebibliography}{53}
\expandafter\ifx\csname natexlab\endcsname\relax\def\natexlab#1{#1}\fi

\bibitem[{Brown et~al.(2020)Brown, Mann, Ryder, Subbiah, Kaplan, Dhariwal, Neelakantan, Shyam, Sastry, Askell et~al.}]{brown2020language}
Tom Brown, Benjamin Mann, Nick Ryder, Melanie Subbiah, Jared~D Kaplan, Prafulla Dhariwal, Arvind Neelakantan, Pranav Shyam, Girish Sastry, Amanda Askell, et~al. 2020.
\newblock Language models are few-shot learners.
\newblock \emph{Advances in neural information processing systems}, 33:1877--1901.

\bibitem[{Budzianowski et~al.(2018)Budzianowski, Wen, Tseng, Casanueva, Ultes, Ramadan, and Ga{\v{s}}i{\'c}}]{budzianowski-etal-2018-multiwoz}
Pawe{\l} Budzianowski, Tsung-Hsien Wen, Bo-Hsiang Tseng, I{\~n}igo Casanueva, Stefan Ultes, Osman Ramadan, and Milica Ga{\v{s}}i{\'c}. 2018.
\newblock \href {https://doi.org/10.18653/v1/D18-1547} {{M}ulti{WOZ} - a large-scale multi-domain {W}izard-of-{O}z dataset for task-oriented dialogue modelling}.
\newblock In \emph{Proceedings of the 2018 Conference on Empirical Methods in Natural Language Processing}, pages 5016--5026, Brussels, Belgium. Association for Computational Linguistics.

\bibitem[{Chaudhry et~al.(2019)Chaudhry, Ranzato, Rohrbach, and Elhoseiny}]{chaudhry2018efficient}
Arslan Chaudhry, Marc'Aurelio Ranzato, Marcus Rohrbach, and Mohamed Elhoseiny. 2019.
\newblock \href {https://openreview.net/forum?id=Hkf2\_sC5FX} {Efficient lifelong learning with {A-GEM}}.
\newblock In \emph{7th International Conference on Learning Representations, {ICLR} 2019, New Orleans, LA, USA, May 6-9, 2019}. OpenReview.net.

\bibitem[{Chen et~al.(2016)Chen, Goodfellow, and Shlens}]{chen2015net2net}
Tianqi Chen, Ian~J. Goodfellow, and Jonathon Shlens. 2016.
\newblock \href {http://arxiv.org/abs/1511.05641} {Net2net: Accelerating learning via knowledge transfer}.
\newblock In \emph{4th International Conference on Learning Representations, {ICLR} 2016, San Juan, Puerto Rico, May 2-4, 2016, Conference Track Proceedings}.

\bibitem[{Chuang et~al.(2020)Chuang, Su, and Chen}]{chuang-etal-2020-lifelong}
Yung-Sung Chuang, Shang-Yu Su, and Yun-Nung Chen. 2020.
\newblock \href {https://doi.org/10.18653/v1/2020.emnlp-main.233} {Lifelong language knowledge distillation}.
\newblock In \emph{Proceedings of the 2020 Conference on Empirical Methods in Natural Language Processing (EMNLP)}, pages 2914--2924, Online. Association for Computational Linguistics.

\bibitem[{Ding et~al.(2023)Ding, Qin, Liu, Chia, Li, Joty, and Bing}]{ding-etal-2023-gpt}
Bosheng Ding, Chengwei Qin, Linlin Liu, Yew~Ken Chia, Boyang Li, Shafiq Joty, and Lidong Bing. 2023.
\newblock \href {https://doi.org/10.18653/v1/2023.acl-long.626} {Is {GPT}-3 a good data annotator?}
\newblock In \emph{Proceedings of the 61st Annual Meeting of the Association for Computational Linguistics (Volume 1: Long Papers)}, pages 11173--11195, Toronto, Canada. Association for Computational Linguistics.

\bibitem[{El-Kassas et~al.(2021)El-Kassas, Salama, Rafea, and Mohamed}]{el2021automatic}
Wafaa~S El-Kassas, Cherif~R Salama, Ahmed~A Rafea, and Hoda~K Mohamed. 2021.
\newblock Automatic text summarization: A comprehensive survey.
\newblock \emph{Expert Systems with Applications}, 165:113679.

\bibitem[{Fernando et~al.(2017)Fernando, Banarse, Blundell, Zwols, Ha, Rusu, Pritzel, and Wierstra}]{fernando2017pathnet}
Chrisantha Fernando, Dylan Banarse, Charles Blundell, Yori Zwols, David Ha, Andrei~A Rusu, Alexander Pritzel, and Daan Wierstra. 2017.
\newblock \href {http://arxiv.org/abs/1701.08734} {Pathnet: Evolution channels gradient descent in super neural networks}.
\newblock \emph{arXiv preprint arXiv:1701.08734}.

\bibitem[{Ham et~al.(2020)Ham, Lee, Jang, and Kim}]{ham-etal-2020-end}
Donghoon Ham, Jeong-Gwan Lee, Youngsoo Jang, and Kee-Eung Kim. 2020.
\newblock \href {https://doi.org/10.18653/v1/2020.acl-main.54} {End-to-end neural pipeline for goal-oriented dialogue systems using {GPT}-2}.
\newblock In \emph{Proceedings of the 58th Annual Meeting of the Association for Computational Linguistics}, pages 583--592, Online. Association for Computational Linguistics.

\bibitem[{Han et~al.(2020)Han, Dai, Gao, Lin, Liu, Li, Sun, and Zhou}]{han-etal-2020-continual}
Xu~Han, Yi~Dai, Tianyu Gao, Yankai Lin, Zhiyuan Liu, Peng Li, Maosong Sun, and Jie Zhou. 2020.
\newblock \href {https://doi.org/10.18653/v1/2020.acl-main.573} {Continual relation learning via episodic memory activation and reconsolidation}.
\newblock In \emph{Proceedings of the 58th Annual Meeting of the Association for Computational Linguistics}, pages 6429--6440, Online. Association for Computational Linguistics.

\bibitem[{He et~al.(2015)He, Lewis, and Zettlemoyer}]{he-etal-2015-question}
Luheng He, Mike Lewis, and Luke Zettlemoyer. 2015.
\newblock \href {https://doi.org/10.18653/v1/D15-1076} {Question-answer driven semantic role labeling: Using natural language to annotate natural language}.
\newblock In \emph{Proceedings of the 2015 Conference on Empirical Methods in Natural Language Processing}, pages 643--653, Lisbon, Portugal. Association for Computational Linguistics.

\bibitem[{Hinton et~al.(2015)Hinton, Vinyals, Dean et~al.}]{hinton2015distilling}
Geoffrey Hinton, Oriol Vinyals, Jeff Dean, et~al. 2015.
\newblock \href {https://arxiv.org/abs/1503.02531} {Distilling the knowledge in a neural network}.
\newblock \emph{arXiv preprint arXiv:1503.02531}, 2(7).

\bibitem[{Houlsby et~al.(2019)Houlsby, Giurgiu, Jastrzebski, Morrone, De~Laroussilhe, Gesmundo, Attariyan, and Gelly}]{houlsby2019parameter}
Neil Houlsby, Andrei Giurgiu, Stanislaw Jastrzebski, Bruna Morrone, Quentin De~Laroussilhe, Andrea Gesmundo, Mona Attariyan, and Sylvain Gelly. 2019.
\newblock \href {https://arxiv.org/abs/1902.00751} {Parameter-efficient transfer learning for nlp}.
\newblock In \emph{International Conference on Machine Learning}, pages 2790--2799. PMLR.

\bibitem[{Hu et~al.(2022)Hu, yelong shen, Wallis, Allen-Zhu, Li, Wang, Wang, and Chen}]{hu2022lora}
Edward~J Hu, yelong shen, Phillip Wallis, Zeyuan Allen-Zhu, Yuanzhi Li, Shean Wang, Lu~Wang, and Weizhu Chen. 2022.
\newblock \href {https://openreview.net/forum?id=nZeVKeeFYf9} {Lo{RA}: Low-rank adaptation of large language models}.
\newblock In \emph{International Conference on Learning Representations}.

\bibitem[{Ke et~al.(2020)Ke, Liu, and Huang}]{ke2020continual}
Zixuan Ke, Bing Liu, and Xingchang Huang. 2020.
\newblock \href {https://proceedings.neurips.cc/paper/2020/file/d7488039246a405baf6a7cbc3613a56f-Paper.pdf} {Continual learning of a mixed sequence of similar and dissimilar tasks}.
\newblock In \emph{Advances in Neural Information Processing Systems}, volume~33, pages 18493--18504. Curran Associates, Inc.

\bibitem[{Ke et~al.(2021)Ke, Liu, Ma, Xu, and Shu}]{ke2021achieving}
Zixuan Ke, Bing Liu, Nianzu Ma, Hu~Xu, and Lei Shu. 2021.
\newblock \href {https://arxiv.org/abs/2112.02706} {Achieving forgetting prevention and knowledge transfer in continual learning}.
\newblock \emph{Advances in Neural Information Processing Systems}, 34:22443--22456.

\bibitem[{Kirkpatrick et~al.(2017)Kirkpatrick, Pascanu, Rabinowitz, Veness, Desjardins, Rusu, Milan, Quan, Ramalho, Grabska-Barwinska et~al.}]{kirkpatrick2017overcoming}
James Kirkpatrick, Razvan Pascanu, Neil Rabinowitz, Joel Veness, Guillaume Desjardins, Andrei~A Rusu, Kieran Milan, John Quan, Tiago Ramalho, Agnieszka Grabska-Barwinska, et~al. 2017.
\newblock \href {https://arxiv.org/abs/1612.00796} {Overcoming catastrophic forgetting in neural networks}.
\newblock \emph{Proceedings of the national academy of sciences}, 114(13):3521--3526.

\bibitem[{Lake et~al.(2017)Lake, Ullman, Tenenbaum, and Gershman}]{lake_ullman_tenenbaum_gershman_2017}
Brenden~M. Lake, Tomer~D. Ullman, Joshua~B. Tenenbaum, and Samuel~J. Gershman. 2017.
\newblock \href {https://doi.org/10.1017/S0140525X16001837} {Building machines that learn and think like people}.
\newblock \emph{Behavioral and Brain Sciences}, 40:e253.

\bibitem[{Li and Liang(2021)}]{li-liang-2021-prefix}
Xiang~Lisa Li and Percy Liang. 2021.
\newblock \href {https://doi.org/10.18653/v1/2021.acl-long.353} {Prefix-tuning: Optimizing continuous prompts for generation}.
\newblock In \emph{Proceedings of the 59th Annual Meeting of the Association for Computational Linguistics and the 11th International Joint Conference on Natural Language Processing (Volume 1: Long Papers)}, pages 4582--4597, Online. Association for Computational Linguistics.

\bibitem[{Li and Hoiem(2017)}]{li2017learning}
Zhizhong Li and Derek Hoiem. 2017.
\newblock \href {http://arxiv.org/abs/1606.09282} {Learning without forgetting}.
\newblock \emph{IEEE transactions on pattern analysis and machine intelligence}, 40(12):2935--2947.

\bibitem[{Lin et~al.(2022{\natexlab{a}})Lin, Yang, Fan, and Zhang}]{lin2022beyond}
Sen Lin, Li~Yang, Deliang Fan, and Junshan Zhang. 2022{\natexlab{a}}.
\newblock \href {https://openreview.net/forum?id=diV1PpaP33} {Beyond not-forgetting: Continual learning with backward knowledge transfer}.
\newblock In \emph{Advances in Neural Information Processing Systems}.

\bibitem[{Lin et~al.(2022{\natexlab{b}})Lin, Yang, Fan, and Zhang}]{lin2022trgp}
Sen Lin, Li~Yang, Deliang Fan, and Junshan Zhang. 2022{\natexlab{b}}.
\newblock \href {https://openreview.net/forum?id=iEvAf8i6JjO} {{TRGP}: Trust region gradient projection for continual learning}.
\newblock In \emph{International Conference on Learning Representations}.

\bibitem[{Liu et~al.(2019)Liu, Simonyan, and Yang}]{liu2018darts}
Hanxiao Liu, Karen Simonyan, and Yiming Yang. 2019.
\newblock \href {https://openreview.net/forum?id=S1eYHoC5FX} {{DARTS}: Differentiable architecture search}.
\newblock In \emph{International Conference on Learning Representations}.

\bibitem[{Lopez-Paz and Ranzato(2017)}]{lopez2017gradient}
David Lopez-Paz and Marc'Aurelio Ranzato. 2017.
\newblock \href {https://arxiv.org/abs/1706.08840} {Gradient episodic memory for continual learning}.
\newblock \emph{Advances in neural information processing systems}, 30.

\bibitem[{Madotto et~al.(2021)Madotto, Lin, Zhou, Moon, Crook, Liu, Yu, Cho, Fung, and Wang}]{madotto-etal-2021-continual}
Andrea Madotto, Zhaojiang Lin, Zhenpeng Zhou, Seungwhan Moon, Paul Crook, Bing Liu, Zhou Yu, Eunjoon Cho, Pascale Fung, and Zhiguang Wang. 2021.
\newblock \href {https://doi.org/10.18653/v1/2021.emnlp-main.590} {Continual learning in task-oriented dialogue systems}.
\newblock In \emph{Proceedings of the 2021 Conference on Empirical Methods in Natural Language Processing}, pages 7452--7467, Online and Punta Cana, Dominican Republic. Association for Computational Linguistics.

\bibitem[{Masoudnia and Ebrahimpour(2014)}]{masoudnia2014mixture}
Saeed Masoudnia and Reza Ebrahimpour. 2014.
\newblock \href {https://doi.org/https://doi.org/10.1007/s10462-012-9338-y} {Mixture of experts: a literature survey}.
\newblock \emph{Artificial Intelligence Review}, 42(2):275--293.

\bibitem[{McCloskey and Cohen(1989)}]{mccloskey1989catastrophic}
Michael McCloskey and Neal~J Cohen. 1989.
\newblock \href {https://doi.org/https://doi.org/10.1016/S0079-7421(08)60536-8} {Catastrophic interference in connectionist networks: The sequential learning problem}.
\newblock In \emph{Psychology of learning and motivation}, volume~24, pages 109--165. Elsevier.

\bibitem[{Mi et~al.(2020)Mi, Chen, Zhao, Huang, and Faltings}]{mi-etal-2020-continual}
Fei Mi, Liangwei Chen, Mengjie Zhao, Minlie Huang, and Boi Faltings. 2020.
\newblock \href {https://doi.org/10.18653/v1/2020.findings-emnlp.310} {Continual learning for natural language generation in task-oriented dialog systems}.
\newblock In \emph{Findings of the Association for Computational Linguistics: EMNLP 2020}, pages 3461--3474, Online. Association for Computational Linguistics.

\bibitem[{Novikova et~al.(2017)Novikova, Du{\v{s}}ek, and Rieser}]{novikova-etal-2017-e2e}
Jekaterina Novikova, Ond{\v{r}}ej Du{\v{s}}ek, and Verena Rieser. 2017.
\newblock \href {https://doi.org/10.18653/v1/W17-5525} {The {E}2{E} dataset: New challenges for end-to-end generation}.
\newblock In \emph{Proceedings of the 18th Annual {SIG}dial Meeting on Discourse and Dialogue}, pages 201--206, Saarbr{\"u}cken, Germany. Association for Computational Linguistics.

\bibitem[{Ouyang et~al.(2022)Ouyang, Wu, Jiang, Almeida, Wainwright, Mishkin, Zhang, Agarwal, Slama, Ray et~al.}]{ouyang2022training}
Long Ouyang, Jeffrey Wu, Xu~Jiang, Diogo Almeida, Carroll Wainwright, Pamela Mishkin, Chong Zhang, Sandhini Agarwal, Katarina Slama, Alex Ray, et~al. 2022.
\newblock Training language models to follow instructions with human feedback.
\newblock \emph{Advances in Neural Information Processing Systems}, 35:27730--27744.

\bibitem[{Pfeiffer et~al.(2020)Pfeiffer, R{\"u}ckl{\'e}, Poth, Kamath, Vuli{\'c}, Ruder, Cho, and Gurevych}]{pfeiffer-etal-2020-adapterhub}
Jonas Pfeiffer, Andreas R{\"u}ckl{\'e}, Clifton Poth, Aishwarya Kamath, Ivan Vuli{\'c}, Sebastian Ruder, Kyunghyun Cho, and Iryna Gurevych. 2020.
\newblock \href {https://doi.org/10.18653/v1/2020.emnlp-demos.7} {{A}dapter{H}ub: A framework for adapting transformers}.
\newblock In \emph{Proceedings of the 2020 Conference on Empirical Methods in Natural Language Processing: System Demonstrations}, pages 46--54, Online. Association for Computational Linguistics.

\bibitem[{Qin and Joty(2022{\natexlab{a}})}]{qin-joty-2022-continual}
Chengwei Qin and Shafiq Joty. 2022{\natexlab{a}}.
\newblock \href {https://doi.org/10.18653/v1/2022.acl-long.198} {Continual few-shot relation learning via embedding space regularization and data augmentation}.
\newblock In \emph{Proceedings of the 60th Annual Meeting of the Association for Computational Linguistics (Volume 1: Long Papers)}, pages 2776--2789, Dublin, Ireland. Association for Computational Linguistics.

\bibitem[{Qin and Joty(2022{\natexlab{b}})}]{qin2022lfpt}
Chengwei Qin and Shafiq Joty. 2022{\natexlab{b}}.
\newblock \href {https://openreview.net/forum?id=HCRVf71PMF} {{LFPT}5: A unified framework for lifelong few-shot language learning based on prompt tuning of t5}.
\newblock In \emph{International Conference on Learning Representations}.

\bibitem[{Qin et~al.(2023{\natexlab{a}})Qin, Joty, Li, and Zhao}]{qin-etal-2023-learning}
Chengwei Qin, Shafiq Joty, Qian Li, and Ruochen Zhao. 2023{\natexlab{a}}.
\newblock \href {https://doi.org/10.18653/v1/2023.acl-long.659} {Learning to initialize: Can meta learning improve cross-task generalization in prompt tuning?}
\newblock In \emph{Proceedings of the 61st Annual Meeting of the Association for Computational Linguistics (Volume 1: Long Papers)}, pages 11802--11832, Toronto, Canada. Association for Computational Linguistics.

\bibitem[{Qin et~al.(2023{\natexlab{b}})Qin, Zhang, Zhang, Chen, Yasunaga, and Yang}]{qin2023chatgpt}
Chengwei Qin, Aston Zhang, Zhuosheng Zhang, Jiaao Chen, Michihiro Yasunaga, and Diyi Yang. 2023{\natexlab{b}}.
\newblock Is chatgpt a general-purpose natural language processing task solver?
\newblock \emph{arXiv preprint arXiv:2302.06476}.

\bibitem[{Radford et~al.(2019)Radford, Wu, Child, Luan, Amodei, Sutskever et~al.}]{radford2019language}
Alec Radford, Jeffrey Wu, Rewon Child, David Luan, Dario Amodei, Ilya Sutskever, et~al. 2019.
\newblock \href {https://d4mucfpksywv.cloudfront.net/better-language-models/language-models.pdf} {Language models are unsupervised multitask learners}.
\newblock \emph{OpenAI blog}, 1(8):9.

\bibitem[{Raffel et~al.(2020)Raffel, Shazeer, Roberts, Lee, Narang, Matena, Zhou, Li, Liu et~al.}]{raffel2020exploring}
Colin Raffel, Noam Shazeer, Adam Roberts, Katherine Lee, Sharan Narang, Michael Matena, Yanqi Zhou, Wei Li, Peter~J Liu, et~al. 2020.
\newblock \href {http://arxiv.org/abs/1910.10683} {Exploring the limits of transfer learning with a unified text-to-text transformer.}
\newblock \emph{J. Mach. Learn. Res.}, 21(140):1--67.

\bibitem[{Rajpurkar et~al.(2016)Rajpurkar, Zhang, Lopyrev, and Liang}]{rajpurkar-etal-2016-squad}
Pranav Rajpurkar, Jian Zhang, Konstantin Lopyrev, and Percy Liang. 2016.
\newblock \href {https://doi.org/10.18653/v1/D16-1264} {{SQ}u{AD}: 100,000+ questions for machine comprehension of text}.
\newblock In \emph{Proceedings of the 2016 Conference on Empirical Methods in Natural Language Processing}, pages 2383--2392, Austin, Texas. Association for Computational Linguistics.

\bibitem[{Rebuffi et~al.(2017)Rebuffi, Kolesnikov, Sperl, and Lampert}]{rebuffi2017icarl}
Sylvestre{-}Alvise Rebuffi, Alexander Kolesnikov, Georg Sperl, and Christoph~H. Lampert. 2017.
\newblock \href {https://doi.org/10.1109/CVPR.2017.587} {icarl: Incremental classifier and representation learning}.
\newblock In \emph{2017 {IEEE} Conference on Computer Vision and Pattern Recognition, {CVPR} 2017, Honolulu, HI, USA, July 21-26, 2017}, pages 5533--5542. {IEEE} Computer Society.

\bibitem[{Ritter et~al.(2018)Ritter, Botev, and Barber}]{ritter2018online}
Hippolyt Ritter, Aleksandar Botev, and David Barber. 2018.
\newblock \href {https://proceedings.neurips.cc/paper/2018/hash/f31b20466ae89669f9741e047487eb37-Abstract.html} {Online structured laplace approximations for overcoming catastrophic forgetting}.
\newblock In \emph{Advances in Neural Information Processing Systems 31: Annual Conference on Neural Information Processing Systems 2018, NeurIPS 2018, December 3-8, 2018, Montr{\'{e}}al, Canada}, pages 3742--3752.

\bibitem[{Ruder et~al.(2019)Ruder, Peters, Swayamdipta, and Wolf}]{ruder-etal-2019-transfer}
Sebastian Ruder, Matthew~E. Peters, Swabha Swayamdipta, and Thomas Wolf. 2019.
\newblock \href {https://doi.org/10.18653/v1/N19-5004} {Transfer learning in natural language processing}.
\newblock In \emph{Proceedings of the 2019 Conference of the North {A}merican Chapter of the Association for Computational Linguistics: Tutorials}, pages 15--18, Minneapolis, Minnesota. Association for Computational Linguistics.

\bibitem[{Rusu et~al.(2016)Rusu, Rabinowitz, Desjardins, Soyer, Kirkpatrick, Kavukcuoglu, Pascanu, and Hadsell}]{rusu2016progressive}
Andrei~A Rusu, Neil~C Rabinowitz, Guillaume Desjardins, Hubert Soyer, James Kirkpatrick, Koray Kavukcuoglu, Razvan Pascanu, and Raia Hadsell. 2016.
\newblock \href {http://arxiv.org/abs/1606.04671} {Progressive neural networks}.
\newblock \emph{arXiv preprint arXiv:1606.04671}.

\bibitem[{See et~al.(2017)See, Liu, and Manning}]{see2017get}
Abigail See, Peter~J Liu, and Christopher~D Manning. 2017.
\newblock \href {https://arxiv.org/abs/1704.04368} {Get to the point: Summarization with pointer-generator networks}.
\newblock \emph{arXiv preprint arXiv:1704.04368}.

\bibitem[{Shin et~al.(2017)Shin, Lee, Kim, and Kim}]{shin2017continual}
Hanul Shin, Jung~Kwon Lee, Jaehong Kim, and Jiwon Kim. 2017.
\newblock \href {https://proceedings.neurips.cc/paper/2017/hash/0efbe98067c6c73dba1250d2beaa81f9-Abstract.html} {Continual learning with deep generative replay}.
\newblock In \emph{Advances in Neural Information Processing Systems 30: Annual Conference on Neural Information Processing Systems 2017, December 4-9, 2017, Long Beach, CA, {USA}}, pages 2990--2999.

\bibitem[{Socher et~al.(2013)Socher, Perelygin, Wu, Chuang, Manning, Ng, and Potts}]{socher-etal-2013-recursive}
Richard Socher, Alex Perelygin, Jean Wu, Jason Chuang, Christopher~D. Manning, Andrew Ng, and Christopher Potts. 2013.
\newblock \href {https://aclanthology.org/D13-1170} {Recursive deep models for semantic compositionality over a sentiment treebank}.
\newblock In \emph{Proceedings of the 2013 Conference on Empirical Methods in Natural Language Processing}, pages 1631--1642, Seattle, Washington, USA. Association for Computational Linguistics.

\bibitem[{Sun et~al.(2020)Sun, Ho, and Lee}]{sun2020lamal}
Fan-Keng Sun, Cheng-Hao Ho, and Hung-Yi Lee. 2020.
\newblock \href {https://openreview.net/forum?id=Skgxcn4YDS} {Lamol: Language modeling for lifelong language learning}.
\newblock In \emph{International Conference on Learning Representations}.

\bibitem[{Wang et~al.(2023)Wang, Fu, Zhang, Zhou, and Zhao}]{wang2023metacognitive}
Han Wang, Ruiliu Fu, Xuejun Zhang, Jun Zhou, and Qingwei Zhao. 2023.
\newblock Metacognitive adaptation to enhance lifelong language learning.
\newblock \emph{IEICE TRANSACTIONS on Information and Systems}, 106(1):86--90.

\bibitem[{Wen et~al.(2015)Wen, Ga{\v{s}}i{\'c}, Mrk{\v{s}}i{\'c}, Su, Vandyke, and Young}]{wen-etal-2015-semantically}
Tsung-Hsien Wen, Milica Ga{\v{s}}i{\'c}, Nikola Mrk{\v{s}}i{\'c}, Pei-Hao Su, David Vandyke, and Steve Young. 2015.
\newblock \href {https://doi.org/10.18653/v1/D15-1199} {Semantically conditioned {LSTM}-based natural language generation for spoken dialogue systems}.
\newblock In \emph{Proceedings of the 2015 Conference on Empirical Methods in Natural Language Processing}, pages 1711--1721, Lisbon, Portugal. Association for Computational Linguistics.

\bibitem[{Wolf et~al.(2020)Wolf, Debut, Sanh, Chaumond, Delangue, Moi, Cistac, Rault, Louf, Funtowicz, Davison, Shleifer, von Platen, Ma, Jernite, Plu, Xu, Le~Scao, Gugger, Drame, Lhoest, and Rush}]{wolf-etal-2020-transformers}
Thomas Wolf, Lysandre Debut, Victor Sanh, Julien Chaumond, Clement Delangue, Anthony Moi, Pierric Cistac, Tim Rault, Remi Louf, Morgan Funtowicz, Joe Davison, Sam Shleifer, Patrick von Platen, Clara Ma, Yacine Jernite, Julien Plu, Canwen Xu, Teven Le~Scao, Sylvain Gugger, Mariama Drame, Quentin Lhoest, and Alexander Rush. 2020.
\newblock \href {https://doi.org/10.18653/v1/2020.emnlp-demos.6} {Transformers: State-of-the-art natural language processing}.
\newblock In \emph{Proceedings of the 2020 Conference on Empirical Methods in Natural Language Processing: System Demonstrations}, pages 38--45, Online. Association for Computational Linguistics.

\bibitem[{Zenke et~al.(2017)Zenke, Poole, and Ganguli}]{zenke2017continual}
Friedemann Zenke, Ben Poole, and Surya Ganguli. 2017.
\newblock \href {http://proceedings.mlr.press/v70/zenke17a.html} {Continual learning through synaptic intelligence}.
\newblock In \emph{Proceedings of the 34th International Conference on Machine Learning, {ICML} 2017, Sydney, NSW, Australia, 6-11 August 2017}, volume~70 of \emph{Proceedings of Machine Learning Research}, pages 3987--3995. {PMLR}.

\bibitem[{Zhang et~al.(2022)Zhang, Wang, and Yang}]{zhang-etal-2022-continual}
Yanzhe Zhang, Xuezhi Wang, and Diyi Yang. 2022.
\newblock \href {https://doi.org/10.18653/v1/2022.acl-long.255} {Continual sequence generation with adaptive compositional modules}.
\newblock In \emph{Proceedings of the 60th Annual Meeting of the Association for Computational Linguistics (Volume 1: Long Papers)}, pages 3653--3667, Dublin, Ireland. Association for Computational Linguistics.

\bibitem[{Zhao et~al.(2023)Zhao, Li, Joty, Qin, and Bing}]{zhao-etal-2023-verify}
Ruochen Zhao, Xingxuan Li, Shafiq Joty, Chengwei Qin, and Lidong Bing. 2023.
\newblock \href {https://doi.org/10.18653/v1/2023.acl-long.320} {Verify-and-edit: A knowledge-enhanced chain-of-thought framework}.
\newblock In \emph{Proceedings of the 61st Annual Meeting of the Association for Computational Linguistics (Volume 1: Long Papers)}, pages 5823--5840, Toronto, Canada. Association for Computational Linguistics.

\bibitem[{Zhong et~al.(2017)Zhong, Xiong, and Socher}]{zhong2017seq2sql}
Victor Zhong, Caiming Xiong, and Richard Socher. 2017.
\newblock \href {https://arxiv.org/abs/1709.00103} {Seq2sql: Generating structured queries from natural language using reinforcement learning}.
\newblock \emph{arXiv preprint arXiv:1709.00103}.

\end{thebibliography}
\bibliographystyle{acl_natbib}

\appendix
\section{Appendix}
\label{sec:appendix}

\begin{figure}[t]
    \centering
    \includegraphics[width=0.48\textwidth]{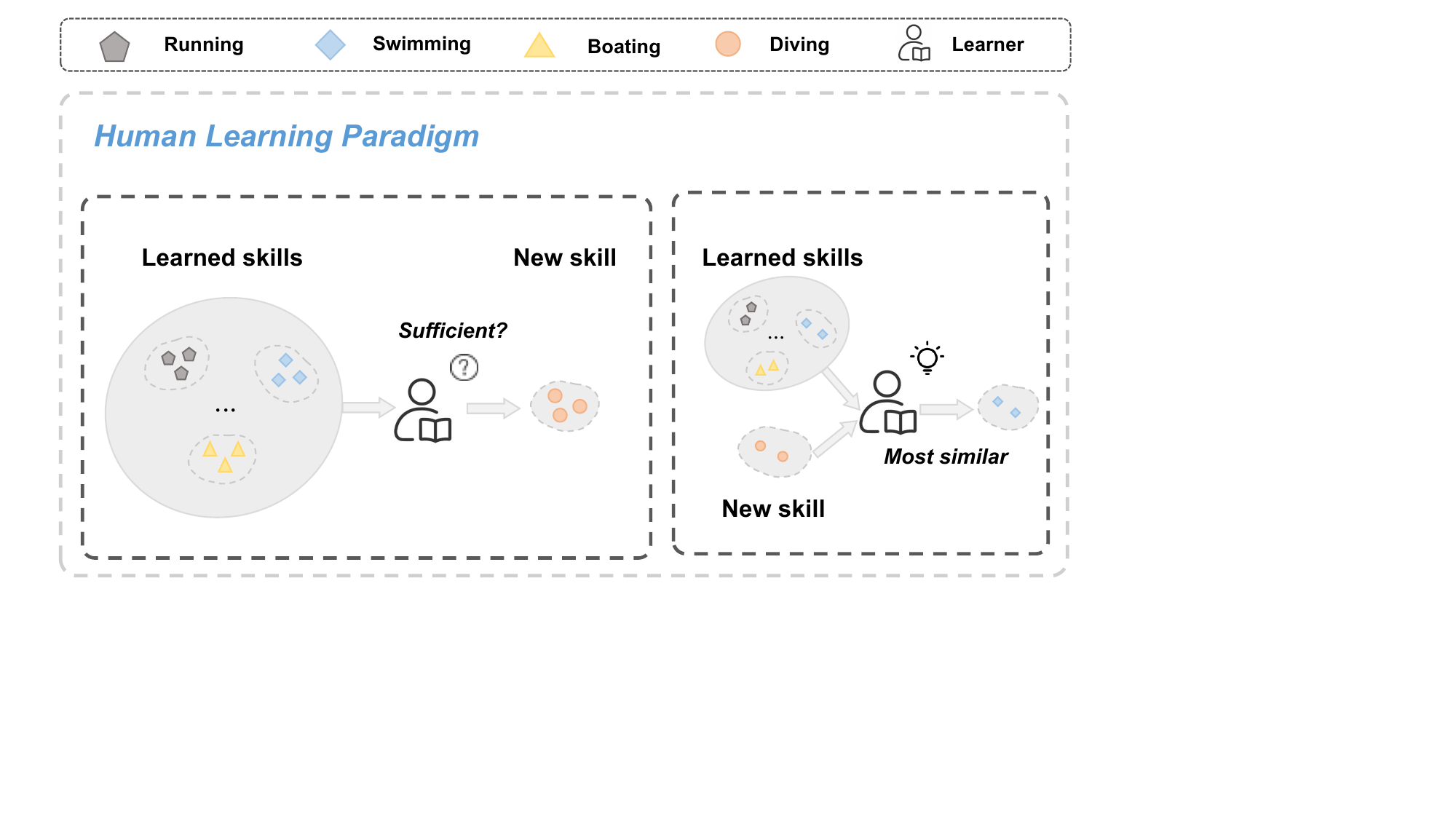}
    \caption{Given three learned skills, \ie\ swimming, running and boating, humans can determine that these skills are \emph{not} sufficient for diving. And after realizing that swimming is the most similar learned skill, they only need to learn the new aspect, \ie\ how to safely jump off the diving platform to master the new diving skill.
    }
    \label{fig:human}
\end{figure}

\begin{figure}[t]
    \centering
    \includegraphics[width=0.48\textwidth]{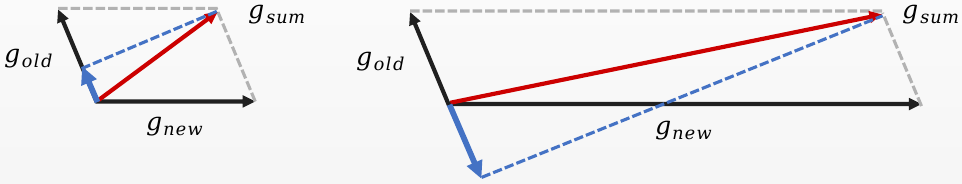}
    \caption{The effect of large gradient norm. $g_{old}$, $g_{new}$, and $g_{sum}$ represent the gradient of replayed tasks, the gradient of the new task, and the aggregated gradient, respectively. The blue arrows show the projection of $g_{sum}$ onto $g_{old}$. If the gradient norm of $g_{new}$ is large (right part of the figure), this projection might deviate too much from $g_{old}$.}
    \label{fig:gradient}
\end{figure}

\begin{figure}[t]
    \centering
    \includegraphics[width=0.48\textwidth]{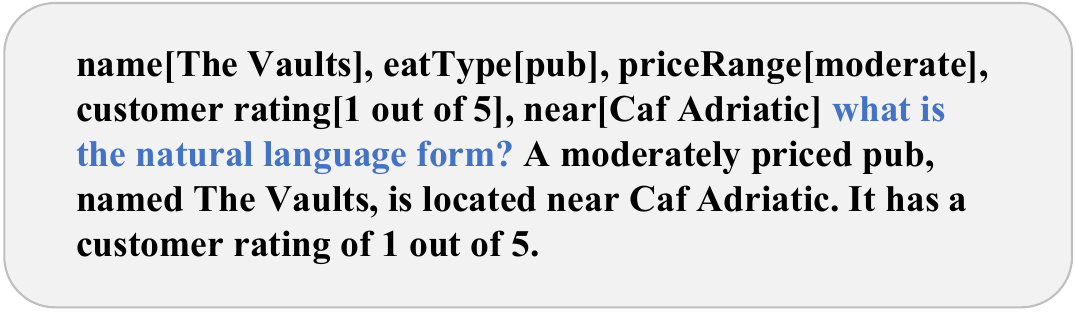}
    \caption{An example of the task instruction for E2ENLG. We color the task instruction in \textcolor{blue}{blue}.}
    \label{fig:e2e_ins}
\end{figure}

\subsection{Illustration of Human Learning} \label{sec:human_learning}
We show the illustration of human learning in \Cref{fig:human}.

\subsection{Effect of Large Gradient Norm} \label{sec:large_gradient}
As shown in the \Cref{fig:gradient}, if the gradient norm of the new task $g_{new}$ is large, the projection of the aggregated gradient $g_{sum}$ onto the gradient of replayed tasks $g_{old}$ might deviate too much from $g_{old}$, leading to more severe forgetting.

\subsection{Task Instruction Example} \label{sec:ins_example}
Following \citet{zhang-etal-2022-continual}, we insert a natural language question describing the purpose of every task (task instruction) after the input of each sample. \Cref{fig:e2e_ins} shows an example of the task instruction for E2ENLG \citep{novikova-etal-2017-e2e}.

\begin{table}[t]
\centering
\small
\begin{tabular}{lll}
\toprule
\multicolumn{2}{l}{\textbf{Order}} & \textbf{Task Sequence} \\
\midrule
\multirow{4}{*}{Similar Tasks}                            & \# 1   & e2e $\shortto$ res $\shortto$ hotel $\shortto$ tv $\shortto$ laptop  \\
                                                          & \# 2   & e2e $\shortto$ tv $\shortto$ res $\shortto$ laptop $\shortto$ hotel   \\
                                                          & \# 3   & res $\shortto$ hotel $\shortto$ e2e $\shortto$ laptop $\shortto$ tv   \\
                                                          & \# 4   & laptop $\shortto$ hotel $\shortto$ res $\shortto$ tv $\shortto$ e2e   \\
\midrule
\multirow{4}{*}{Random Tasks}                            & \# 1   & mwoz $\shortto$ cnn $\shortto$ e2e $\shortto$ res $\shortto$ hotel   \\
                                                         & \# 2   & e2e $\shortto$ sql $\shortto$ hotel $\shortto$ mwoz $\shortto$ res   \\
                                                         & \# 3   & cnn $\shortto$ hotel $\shortto$ sql $\shortto$ e2e $\shortto$ mwoz   \\
                                                         & \# 4   & e2e $\shortto$ mwoz $\shortto$ laptop $\shortto$ sql $\shortto$ tv  \\ \bottomrule
\end{tabular}

\caption{\label{table-task-order}
Different task orders for each scenario. `e2e' stands for E2ENLG. `res', `hotel', `laptop' and `tv' are four domains in RNNLG (restaurant, hotel, laptop and television). `sql', `cnn' and `mwoz' respectively stand for `WikiSQL' (SQL query generation), `CNNDM' (summarization) and `MultiWOZ' (task-oriented dialogue).
}
\end{table}

\begin{table}[t]
    \centering
    \begin{tabular}{ll}
    \toprule
    \textbf{Dataset}       & \textbf{Metric}               \\ \midrule
     RNNLG         & \multirow{3}{*}{A-RG}  \\
     E2ENLG        &                        \\ 
     CNNDM         &                        \\ \hline
     WikiSQL       & lfEM                   \\
     MultiWOZ      & dsEM                   \\ \bottomrule
    \end{tabular}
    \caption{\label{tab:task-specific-metrics}
        Details of task-specific evaluation metrics. `A-RG', `lfEM' and `dsEM' respectively stand for `average of ROUGE-1, ROUGE-2 and ROUGE-L scores', `exact match of logical forms' and `exact match of dialogue state'.
    }
\end{table}

\subsection{Task Orders} \label{sec:diff_task_orders}
We present different task orders for two \lsg\ scenarios in \Cref{table-task-order}.

\subsection{Task-specific Evaluation Metrics} \label{sec:task_metrics}
We report details of task-specific evaluation metrics in \Cref{tab:task-specific-metrics}.

\subsection{Implementation Details} \label{sec:hyper}

All methods are implemented with PyTorch/Transformers library \citep{wolf-etal-2020-transformers}. We adopt AdapterHub \citep{pfeiffer-etal-2020-adapterhub} to implement adapter modules. For hyperparameters, we mainly follow the settings in \citet{zhang-etal-2022-continual} to have a fair comparison. In the expansion stage, we train the model for 6 epochs before selecting modules. In the adaptation stage, we set the number ($n$) of samples selected to obtain the input subspace as $100$. The threshold $\epsilon$ is set as $0.95$ for selecting left-singular vectors. We adopt $1$ for the number of similar tasks $K$. For dynamic gradient scaling, we set $100$ for the number ($q$) of examples selected to calculate the gradient.

\begin{table}[t]
\centering
\small
\setlength\tabcolsep{3pt}
\begin{tabular}{l|ccccc|c}
\toprule
\textbf{Similar \#1} & e2e & res & hotel & tv & laptop & \textbf{Avg} \\
\midrule
ACM & 49.6 & 65.7  & 65.8 & 71.6 & 71.1 &  64.8 \\
\dmea & 49.2 & 67.1  & 68.1  & 72.5 & 72.0 & 65.8 \\

\toprule
\textbf{Similar \#2} & e2e & tv & res & laptop & hotel & \textbf{Avg} \\
\midrule
ACM & 48.7  & 74.0 & 64.4 & 72.6 & 65.0 & 64.9 \\
\dmea & 47.9 & 74.9 & 64.9 & 74.2 & 65.9 & 65.6 \\

\toprule
\textbf{Similar \#3} & res & hotel & e2e & laptop & tv & \textbf{Avg} \\
\midrule
ACM & 65.6 & 67.3 & 48.5 & 72.0 & 69.3 & 64.5 \\
\dmea & 66.9 & 66.6 & 49.5 & 73.7 & 70.9 & 65.5 \\

\toprule
\textbf{Similar \#4} & laptop & hotel & res & tv & e2e & \textbf{Avg} \\
\midrule
ACM & 73.1 & 66.8 & 66.9 & 72.4 & 47.6 & 65.4 \\
\dmea & 74.6 & 67.6 & 67.4 & 72.9 & 48.3 & 66.2 \\

\toprule
\textbf{Random \#1} & mwoz & cnn & e2e & res & hotel & \textbf{Avg} \\
\midrule
ACM & 81.6  & 26.0 & 47.5 & 64.5 & 64.1 & 56.7 \\
\dmea & 81.6 & 26.5 & 48.1 & 65.7 & 65.4 & 57.5 \\

\toprule
\textbf{Random \#2} & e2e & sql & hotel & mwoz & res & \textbf{Avg} \\
\midrule
ACM & 48.4 & 62.7 & 64.6 & 84.8 & 64.0 & 64.9 \\
\dmea & 48.7 & 64.9 & 64.9 & 84.9 & 64.6 & 65.6 \\

\toprule
\textbf{Random \#3} & cnn & hotel & sql & e2e & mwoz & \textbf{Avg} \\
\midrule
ACM & 26.3 & 63.2 & 62.1 & 47.7 & 82.4 & 56.3 \\
\dmea & 26.6 & 65.0 & 63.1 & 48.4 & 83.5 & 57.3 \\

\toprule
\textbf{Random \#4} & e2e & mwoz & laptop & sql & tv & \textbf{Avg} \\
\midrule
ACM & 48.6 & 80.5 & 70.3 & 63.5 & 68.2 & 66.2 \\
\dmea & 49.0 & 82.8 & 71.7 & 64.4 & 68.9 & 67.4 \\

\bottomrule
\end{tabular}
\caption{The performance of each task for every sequence after learning all tasks.}
\label{tab:each_performance}
\end{table}

\begin{table}[t]
\centering
\small
\setlength\tabcolsep{3pt}
\scalebox{0.90}{
\begin{tabular}{l|c|c}
\toprule
\multirow{1}{*}{\textbf{Method}} & \multicolumn{1}{c}{Avg Para Num}                    & \multicolumn{1}{|c}{Avg Time (min)}                    \\
\midrule

ACM & 4.6M & 218.1 \\
\dmea & 4.8M & 223.5 \\
\bottomrule
\end{tabular}
}
\caption{ The comparison of the average number of learnable parameters (Avg Para Num) and average running time (Avg Time) between ACM and \dmea.} 
\label{num_para_running_time}
\end{table}

\subsection{Performance of Each Task} \label{sec:detail_performance}
\Cref{tab:each_performance} shows the performance of each task for every task sequence after learning all tasks. 

\begin{table*}[t]
\centering
\scalebox{0.82}{\begin{tabular}{ll}
\toprule
\multicolumn{2}{l}{\textbf{RNNLG.hotel:} inform(name='mandarin oriental san francisco';dogsallowed='yes';pricerange='ultra high end')}                                                                                    \\ \midrule
\multicolumn{1}{l}{Reference}     & \textit{the mandarin oriental san francisco is in the ultra high end price range and allows dogs.}\\ \hline
\multicolumn{1}{l}{ACM} & \textit{the mandarin oriental san francisco is \blue{a hotel} in the ultra high end range \textcolor{red}{(missing: and allows dogs}).}             \\ \hline
\multicolumn{1}{l}{\dmea}          & \textit{the mandarin oriental san francisco offers ultra high end accommodations and allows dogs.}                                                                                   \\ \toprule
\multicolumn{2}{l}{\textbf{WikiSQL:} on which date was the winning driver alain prost and had damon hill in the pole position ?}                                                                                                                          \\ \midrule
\multicolumn{1}{l}{Reference}     & \textit{select date from table where winning driver = alain prost and pole position = damon hill}                                                                                                                          \\ \hline
\multicolumn{1}{l}{ACM}    & \textit{select date from table where winning driver = alain prost and pole position = damon \textcolor{red}{(missing: hill)}}                                                                                                                     \\ \hline
\multicolumn{1}{l}{\dmea}          & \textit{select date from table where pole position = damon hill and winning driver = alain prost}                                                                                                                    \\ \bottomrule
\end{tabular}}
\caption{Output examples of different methods after learning the whole sequence. We color missing/wrong information in \textcolor{red}{red} and redundant information in \blue{blue}.}
\label{table:case-study-example}
\end{table*}

\begin{table}[t]
\centering
\small
\setlength\tabcolsep{3pt}
\scalebox{0.90}{
\begin{tabular}{l|c|c}
\toprule
\multirow{1}{*}{\textbf{Method}} & \multicolumn{1}{c}{Similar Tasks}                    & \multicolumn{1}{|c}{Random Tasks}                    \\
\midrule

ACM & 64.5 & 66.2 \\
ACM \emph{w} DI & \textbf{64.7}  & \textbf{66.5} \\
\bottomrule
\end{tabular}
}
\caption{ The performance comparison between ACM and ACM with dynamic initialization (ACM \emph{w} DI).} 
\label{acm_with_dynamic_initialization}
\end{table}

\subsection{Number of Learnable Parameters and Computational Resources} \label{sec:comparison_acm_dmea}
We present the average number of learnable parameters and average running time for ACM and \dmea\ in \Cref{num_para_running_time}. From the comparison, we can observe that \dmea\ can outperform ACM with a negligible increase in learnable parameters and computational resources.

\subsection{Learned Model Architecture} \label{sec:learned_model_architecture}

To further demonstrate that dynamically initializing learnable coefficients can facilitate finding the optimal model architecture, we analyze the model expansion stage of ACM and \dmea\ using sequence \#4 in \emph{random} scenario. For the final task \emph{tv}, ACM decides to reuse modules from the first (\emph{e2e}) and the third task (\emph{laptop}) while \dmea\ reuses all modules from \emph{laptop} which is consistent with the observation that the similarity between \emph{tv} and \emph{laptop} is much higher than that between \emph{tv} and \emph{e2e}.

\subsection{Case Study of Model Output} \label{sec:case_study_output}
We select RNNLG.hotel (sequence \#1 in \emph{similar} scenario) and WikiSQL (sequence \#4 in \emph{random} scenario) as two representative tasks and show several examples of output in \Cref{table:case-study-example}. Compared with ACM, \dmea\ possesses the capability to convey more precise and relevant information from the input without introducing superfluous details.

\subsection{Case Study of Dynamic Gradient Scaling} \label{sec:case_gradient_scaling}
The ablation study in \Cref{subsec:abl} demonstrates the importance of dynamic gradient scaling. We further conduct a case study using sequence \#1 in \emph{random} scenario. During the learning of this sequence, the fourth task \emph{res} reuses several modules from the third task \emph{e2e}. After applying dynamic gradient scaling, the performance of \emph{e2e} is improved by $0.3$ without compromising \emph{res}, indicating that it does mitigate the bias towards the new task.

\subsection{Case Study of Task Selection} \label{sec:case_selected_similar_task}
To verify that the previous task chosen in the selection stage is indeed the most similar to the new task, we analyze several cases using sequence \#2 in \emph{random} scenario. For the third task \emph{hotel}, the selected first task \emph{e2e} has the highest similarity score as they share the same task type. In addition, the third task \emph{hotel} shares a similar semantic space with the final task \emph{res}. Therefore, it is selected for forward knowledge transfer when learning \emph{res}.

\subsection{Generalization of Dynamic Initialization} \label{sec:general_dynamic_initialization}
To demonstrate the generalization ability of dynamic initialization, we apply it to the expansion stage of ACM. For each scenario, we randomly pick one sequence for experiments. As reported in \Cref{acm_with_dynamic_initialization}, dynamic initialization does benefit ACM, verifying its generalization capability.

\subsection{Real World Application} \label{sec:real_world_application}
Apart from the aforementioned sequence generation tasks, \dmea\ demonstrates the potential to be applied to various real-world lifelong learning scenarios. For example, it can continually train a model to perform summarization and question-answering based on news articles from different domains during the onset of an emerging event like Covid-19.

\subsection{Hyperparameter Search} \label{sec:hyper_search}

We select the number of training epochs before modules selection from $\{6,9,12\}$, the number ($n$) of samples picked to obtain the input subspace from $\{50,100,200,500\}$ and the threshold $\epsilon$ for selecting left-singular vectors from $\{0.90,0.95,0.99\}$. The number of similar previous tasks $K$ is selected from $\{1,2,3\}$. The number ($q$) of examples for calculating the gradient in dynamic gradient scaling is selected from $\{20,50,100,200\}$.

\end{document}